# A Temporal Description Logic for Reasoning about Actions and Plans

**Alessandro Artale**                                                                ARTALE@IRST.ITC.IT
*ITC-IRST, Cognitive and Communication Technologies Division*
*I-38050 Povo TN, Italy*

**Enrico Franconi**                                                                FRANCONI@CS.MAN.AC.UK
*Department of Computer Science, University of Manchester*
*Manchester M13 9PL, UK*

## Abstract

A class of interval-based temporal languages for uniformly representing and reasoning about actions and plans is presented. Actions are represented by describing what is true while the action itself is occurring, and plans are constructed by temporally relating actions and world states. The temporal languages are members of the family of Description Logics, which are characterized by high expressivity combined with good computational properties. The subsumption problem for a class of temporal Description Logics is investigated and sound and complete decision procedures are given. The basic language $\mathcal{TL}$-$\mathcal{F}$ is considered first: it is the composition of a temporal logic $\mathcal{TL}$ – able to express interval temporal networks – together with the non-temporal logic $\mathcal{F}$ – a Feature Description Logic. It is proven that subsumption in this language is an NP-complete problem. Then it is shown how to reason with the more expressive languages $\mathcal{TLU}$-$\mathcal{FU}$ and $\mathcal{TL}$-$\mathcal{ALCF}$. The former adds disjunction both at the temporal and non-temporal sides of the language, the latter extends the non-temporal side with set-valued features (i.e., roles) and a propositionally complete language.

## 1. Introduction

The representation of temporal knowledge has received considerable attention in the Artificial Intelligence community in an attempt to extend existing knowledge representation systems to deal with actions and change. At the same time, many logic-based formalisms were developed and analyzed by logicians and philosophers for the same purposes. In this class of logical formalisms, properties such as expressive power and computability have been studied as regards typical problems involving events and actions.

This paper analyzes from a theoretical point of view the logical and computational properties of a knowledge representation system that allows us to deal with time, actions and plans in a uniform way. The most common approaches to model actions are based on the notion of *state change* – e.g., the formal models based on the original *situation calculus* (McCarthy & Hayes, 1969; Sandewall & Shoham, 1994) or the STRIPS-like planning systems (Fikes & Nilsson, 1971; Lifschitz, 1987) – in which actions are generally considered instantaneous and defined as functions from one state to another by means of pre- and post-conditions. Here, an explicit notion of time is introduced in the modeling language and actions are defined as *occurring over time intervals*, following the Allen proposal (Allen, 1991). In this formalism an action is represented by describing the time course of the world while the action occurs. Concurrent or overlapping actions are allowed: effects of overlapping actions can be different from the sum of their individual effects; effects may not directly follow the action but more complex temporal relations may hold. For instance, consider the motion of a pointer on a screen driven by a mouse: the pointer moves because of the movement of the device on the pad – there is a cause-effect relation – but the two events are contemporary, in the common-sense notion of the word.

A class of interval temporal logics is studied based on Description Logics and inspired by the works of Schmiedel (1990) and of Weida and Litman (1992). In this class of formalisms a *state* describes a collection of properties of the world holding at a certain time. *Actions* are represented through temporal constraints on world states, which pertain to the action itself. *Plans* are built by temporally relating actions and states. To represent the temporal dimension classical Description Logics are extended with temporal constructors; thus a uniform representation for states, actions and plans is provided. Furthermore, the distinction made by Description Logics between the terminological and assertional aspects of the knowledge allows us to describe actions and plans both at an abstract level (action/plan types) and at an instance level (individual actions and plans). In this environment, the *subsumption* calculus is the main inference tool for managing collections of action and plan types. Action and plan types can be organized in a subsumption-based taxonomy, which plays the role of an action/plan library to be used for the tasks known in the literature as plan retrieval and individual plan recognition (Kautz, 1991). A refinement of the *plan recognition* notion is proposed, by splitting it into the different tasks of *plan description classification* – involving a plan type – and *specific plan recognition with respect to a plan description* – involving an individual plan. According to the latter reasoning task, the system is able to recognize which type of action/plan has taken place at a certain time interval, given a set of observations of the world.

Advantages of using Description Logics are their high expressivity combined with desirable computational properties – such as decidability, soundness and completeness of deduction procedures (Buchheit, Donini, & Schaerf, 1993; Schaerf, 1994; Donini, Lenzerini, Nardi, & Schaerf, 1994; Donini, Lenzerini, Nardi, & Nutt, 1995). The main purpose of this work is to investigate a class of *decidable* temporal Description Logics, and to provide complete algorithms for computing subsumption. To this aim, we start with $\mathcal{TL}\text{-}\mathcal{F}$, a language being the composition of a temporal logic $\mathcal{TL}$ – able to express interval temporal networks – together with the non-temporal Description Logic $\mathcal{F}$ – a Feature Description Logic (Smolka, 1992). It turns out that subsumption for $\mathcal{TL}\text{-}\mathcal{F}$ is an NP-complete problem. Then, we show how to reason with more expressive languages: $\mathcal{TLU}\text{-}\mathcal{FU}$, which adds disjunction both at the temporal and non-temporal sides of the language, and $\mathcal{TL}\text{-}\mathcal{ALCF}$, which extends the non-temporal side with set-valued features (i.e., roles) and a propositionally complete Description Logic (Hollunder & Nutt, 1990). In both cases we show that reasoning is decidable and we supply sound and complete procedures for computing subsumption.

The paper is organized as follows. After introducing the main features of Description Logics in Section 2, Section 3 organizes the intuitions underlying our proposal. The technical bases are introduced by briefly overviewing the temporal extensions of Description Logics relevant for this approach – together with the inter-relationships with the interval temporal modal logic – specifically intended for time and action representation and reasoning. The basic feature temporal language ($\mathcal{TL}$-$\mathcal{F}$) is introduced in Section 4. The language syntax is first described in Section 4.1, together with a worked out example illustrating the informal meaning of temporal expressions. Section 4.2 reveals the model theoretic semantics of $\mathcal{TL}$-$\mathcal{F}$, together with a formal definition of the subsumption and instance recognition problems. Section 5 shows that the temporal language is suitable for action and plan representation and reasoning: the well known *cooking* domain and *blocks world* domain are taken into consideration. The sound and complete calculus for the feature temporal language $\mathcal{TL}$-$\mathcal{F}$ is presented in details in Section 6. A proof that subsumption for $\mathcal{TL}$-$\mathcal{F}$ is an NP-complete problem is included. The calculus for $\mathcal{TL}$-$\mathcal{F}$ forms the basic reasoning procedure that can be adapted to deal with logics having an extended propositional part. An algorithm for checking subsumption in presence of disjunction ($\mathcal{TLU}$-$\mathcal{FU}$) is devised in Section 7.1; while in Section 7.2 the non-temporal part of the language is extended with roles and full propositional calculus ($\mathcal{TL}$-$\mathcal{ALCF}$). In both cases, the subsumption problem is still decidable. Operators for homogeneity and persistence are presented in Section 8 for an adequate representation of world states. In particular, a possible solution to the *frame problem*, i.e., the problem to compute what remains unchanged by an action, is suggested. Section 9 surveys the whole spectrum of extensions of Description Logics for representing and reasoning with time and action. This Section is concluded by a comparison with *State Change* based approaches by briefly illustrating the effort made in the situation calculus area to temporally extend this class of formalisms. Section 10 concludes the paper.

## 2. Description Logics

Description Logics[1] are formalisms designed for a logical reconstruction of representation tools such as *frames, semantic networks, object-oriented* and *semantic* data models – see (Calvanese, Lenzerini, & Nardi, 1994) for a survey. Nowadays, Description Logics are also considered the most important unifying formalism for the many object-centered representation languages used in areas other than Knowledge Representation. Important characteristics of Description Logics are high expressivity together with decidability, which guarantee the existence of reasoning algorithms that always terminate with the correct answers.

This Section gives a brief introduction to a basic Description Logic, which will serve as the basic representation language for our proposal. As for the formal apparatus, the formalism introduced by (Schmidt-Schauß & Smolka, 1991) and further elaborated by (Donini, Hollunder, Lenzerini, Spaccamela, Nardi, & Nutt, 1992; Donini et al., 1994, 1995; Buchheit et al., 1993; De Giacomo & Lenzerini, 1995, 1996) is followed: in this way, Description Logics are considered as a *structured* fragment of predicate logic. $\mathcal{ALC}$ (Schmidt-Schauß & Smolka, 1991) is the minimal Description Logic including full negation and disjunction – i.e., propositional calculus, and it is a notational variant of the propositional modal logic $K_{(m)}$ (Halpern & Moses, 1985; Schild, 1991).

The basic types of a Description Logic are *concepts*, *roles*, *features*, and *individuals*. A concept is a description gathering the common properties among a collection of individuals; from a logical point of view it is a unary predicate ranging over the domain of individu-

---

1. Description Logics have been also called *Frame-Based Description Languages, Term Subsumption Languages, Terminological Logics, Taxonomic Logics, Concept Languages* or KL-ONE-*like languages*.

$$
\begin{aligned}
C, D \;\to\; & A \mid && \text{(atomic concept)} \\
& \top \mid && \text{(top)} \\
& \bot \mid && \text{(bottom)} \\
& \neg C \mid && \text{(complement)} \\
& C \sqcap D \mid && \text{(conjunction)} \\
& C \sqcup D \mid && \text{(disjunction)} \\
& \forall P.C \mid && \text{(universal quantifier)} \\
& \exists P.C \mid && \text{(existential quantifier)} \\
& p : C \mid && \text{(selection)} \\
& p \downarrow q \mid && \text{(agreement)} \\
& p \uparrow q \mid && \text{(disagreement)} \\
& p \uparrow && \text{(undefinedness)} \\
p, q \;\to\; & f \mid && \text{(atomic feature)} \\
& p \circ q && \text{(path)}
\end{aligned}
$$

Figure 1: Syntax rules for the $\mathcal{ALCF}$ Description Logic.

als. Properties are represented either by means of roles – which are interpreted as binary relations associating to individuals of a given class values for that property – or by means of features – which are interpreted as functions associating to individuals of a given class a *single* value for that property. The language $\mathcal{ALCF}$, extending $\mathcal{ALC}$ with features (i.e., functional roles) is considered. By the syntax rules of Figure 1, $\mathcal{ALCF}$ concepts (denoted by the letters $C$ and $D$) are built out of *atomic concepts* (denoted by the letter $A$), *atomic roles* (denoted by the letter $P$), and *atomic features* (denoted by the letter $f$). The syntax rules are expressed following the tradition of Description Logics (Baader, Bürckert, Heinsohn, Hollunder, Müller, Nebel, Nutt, & Profitlich, 1990).

The *meaning* of concept expressions is defined as sets of individuals, as for unary predicates, and the meaning of roles as sets of pairs of individuals, as for binary predicates. Formally, an *interpretation* is a pair $\mathcal{I} = (\Delta^{\mathcal{I}}, \cdot^{\mathcal{I}})$ consisting of a set $\Delta^{\mathcal{I}}$ of individuals (the *domain* of $\mathcal{I}$) and a function $\cdot^{\mathcal{I}}$ (the *interpretation function* of $\mathcal{I}$) mapping every concept to a subset of $\Delta^{\mathcal{I}}$, every role to a subset of $\Delta^{\mathcal{I}} \times \Delta^{\mathcal{I}}$, every feature to a partial function from $\Delta^{\mathcal{I}}$ to $\Delta^{\mathcal{I}}$, and every individual into a different element of $\Delta^{\mathcal{I}}$ – i.e., $a^{\mathcal{I}} \neq b^{\mathcal{I}}$ if $a \neq b$ (Unique Name Assumption) – such that the equations of the left column in Figure 2 are satisfied.

The $\mathcal{ALCF}$ semantics identifies concept expressions as fragments of first-order predicate logic. Since the interpretation $\mathcal{I}$ assigns to every atomic concept, role or feature a unary or binary (functional) relation over $\Delta^{\mathcal{I}}$, respectively, one can think of atomic concepts, roles and features as unary and binary (functional) predicates. This can be seen as follows: an atomic concept $A$, an atomic role $P$, and an atomic feature $f$, are mapped respectively to the open formulas $F_A(\gamma)$, $P(\alpha, \beta)$, and $F_f(\alpha, \beta)$ with $F_f$ satisfying the functionality axiom $\forall y, z. F_f(x, y) \land F_f(x, z) \to y = z$ – i.e., $F_f$ is a functional relation.

The rightmost column of Figure 2 gives the transformational semantics of $\mathcal{ALCF}$ expressions in terms of FOL well-formed formulæ, while the left column gives the standard extensional semantics. As far as the transformational semantics is concerned, a concept $C$, a role $P$ and a path $p$ correspond to the FOL open formulae $F_C(\gamma)$, $F_P(\alpha, \beta)$, and $F_p(\alpha, \beta)$,

$$
\begin{array}{rll}
\top^{\mathcal{I}} = & \Delta^{\mathcal{I}} & \text{true} \\
\bot^{\mathcal{I}} = & \emptyset & \text{false} \\
(\neg C)^{\mathcal{I}} = & \Delta^{\mathcal{I}} \setminus C^{\mathcal{I}} & \neg F_C(\gamma) \\
(C \sqcap D)^{\mathcal{I}} = & C^{\mathcal{I}} \cap D^{\mathcal{I}} & F_C(\gamma) \wedge F_D(\gamma) \\
(C \sqcup D)^{\mathcal{I}} = & C^{\mathcal{I}} \cup D^{\mathcal{I}} & F_C(\gamma) \vee F_D(\gamma) \\
(\exists P.C)^{\mathcal{I}} = & \{a \in \Delta^{\mathcal{I}} \mid \exists b.(a,b) \in P^{\mathcal{I}} \wedge b \in C^{\mathcal{I}}\} & \exists x. F_P(\gamma, x) \wedge F_C(x) \\
(\forall P.C)^{\mathcal{I}} = & \{a \in \Delta^{\mathcal{I}} \mid \forall b.(a,b) \in P^{\mathcal{I}} \Rightarrow b \in C^{\mathcal{I}}\} & \forall x. F_P(\gamma, x) \Rightarrow F_C(x) \\
(p:C)^{\mathcal{I}} = & \{a \in dom\, p^{\mathcal{I}} \mid p^{\mathcal{I}}(a) \in C^{\mathcal{I}}\} & \exists x. F_p(\gamma, x) \wedge F_C(x) \\
p \downarrow q^{\mathcal{I}} = & \{a \in dom\, p^{\mathcal{I}} \cap dom\, q^{\mathcal{I}} \mid p^{\mathcal{I}}(a) = q^{\mathcal{I}}(a)\} & \left(\exists x. F_p(\gamma, x) \wedge F_q(\gamma, x)\right) \\
p \uparrow q^{\mathcal{I}} = & \{a \in dom\, p^{\mathcal{I}} \cap dom\, q^{\mathcal{I}} \mid p^{\mathcal{I}}(a) \ne q^{\mathcal{I}}(a)\} & \left(\exists x,y. F_p(\gamma, x) \wedge F_q(\gamma, y)\right) \wedge \\
& & \left(\forall x,y. F_p(\gamma, x) \wedge F_q(\gamma, y) \to x \ne y\right) \\
(p\uparrow)^{\mathcal{I}} = & \Delta^{\mathcal{I}} \setminus dom\, p^{\mathcal{I}} & \neg \exists x. F_p(\gamma, x) \\
(p \circ q)^{\mathcal{I}} = & p^{\mathcal{I}} \circ q^{\mathcal{I}} & \exists x. F_p(\alpha, x) \wedge F_q(x, \beta)
\end{array}
$$

Figure 2: The extensional and transformational semantics in $\mathcal{ALCF}$.

respectively. It is worth noting that the extensional semantics of the left column gives also an interpretation for the formulas of the right column so that the following proposition holds.

**Proposition 2.1 (Concepts vs. fol formulæ)** *Let $C$ be an $\mathcal{ALCF}$ concept expression. Then the transformational semantics of Figure 2 maps $C$ into a logically equivalent first order formula.*

A *terminology* or *TBox* is a finite set of *terminological axioms*. For an atomic concept $A$, called *defined concept*, and a (possibly complex) concept $C$, a terminological axiom is of the form $A \doteq C$. An atomic concept not appearing on the left-hand side of any terminological axiom is called a *primitive concept*. *Acyclic simple* TBoxes only are considered: a defined concept may appear at most once as the left-hand side of an axiom, and no terminological cycles are allowed, i.e., no defined concept may occur – neither directly nor indirectly – within its own definition (Nebel, 1991). An interpretation $\mathcal{I}$ *satisfies* $A \doteq C$ if and only if $A^{\mathcal{I}} = C^{\mathcal{I}}$.

As an example, consider the unary relation (i.e., a concept) denoting the class of happy fathers, defined using the atomic predicates Man, Doctor, Rich, Famous (concepts) and CHILD, FRIEND (roles):

HappyFather $\doteq$ Man $\sqcap$ ($\exists$CHILD.$\top$) $\sqcap$ $\forall$CHILD.(Doctor $\sqcap$ $\exists$FRIEND.(Rich $\sqcup$ Famous))

i.e., the men whose children are doctors having some rich or famous friend.

An *ABox* is a finite set of *assertional axioms*, i.e. predications on individual objects. Let $\mathcal{O}$ be the alphabet of symbols denoting *individuals*; an assertion is an axiom of the form $C(a)$, $R(a,b)$ or $p(a,b)$, where $a$ and $b$ denote individuals in $\mathcal{O}$. $C(a)$ is satisfied by an interpretation $\mathcal{I}$ iff $a^{\mathcal{I}} \in C^{\mathcal{I}}$, $P(a,b)$ is satisfied by $\mathcal{I}$ iff $(a^{\mathcal{I}}, b^{\mathcal{I}}) \in P^{\mathcal{I}}$, and $p(a,b)$ is satisfied by $\mathcal{I}$ iff $p^{\mathcal{I}}(a^{\mathcal{I}}) = b^{\mathcal{I}}$.

A *knowledge base* is a finite set $\Sigma$ of terminological and assertional axioms. An interpretation $\mathcal{I}$ is a *model* of a knowledge base $\Sigma$ iff every axiom of $\Sigma$ is satisfied by $\mathcal{I}$. $\Sigma$ *logically implies* $A \sqsubseteq C$ (written $\Sigma \models A \sqsubseteq C$) if $A^{\mathcal{I}} \subseteq C^{\mathcal{I}}$ for every model of $\Sigma$: we say that $A$ is *subsumed* by $C$ in $\Sigma$. The reasoning problem of checking whether $A$ is *subsumed* by $C$ in $\Sigma$ is called *subsumption checking*. $\Sigma$ *logically implies* $C(a)$ (written $\Sigma \models C(a)$) if $a^{\mathcal{I}} \in C^{\mathcal{I}}$ for every model of $\Sigma$: we say that $a$ is an *instance* of $C$ in $\Sigma$. The reasoning problem of checking whether $a$ is an *instance* of $C$ in $\Sigma$ is called *instance recognition*.

An acyclic simple TBox can be transformed into an *expanded* TBox having the same models, where no defined concept makes use in its definition of any other defined concept. In this way, the interpretation of a defined concept in an expanded TBox does not depend on any other defined concept. It is easy to see that $A$ is subsumed by $C$ in an acyclic simple TBox $\Sigma$ if and only if the expansion of $A$ with respect to $\Sigma$ is subsumed by the expansion of $C$ with respect to $\Sigma$ in the empty TBox. The expansion procedure recursively substitutes every defined concept occurring in a definition with its defining expression; such a procedure may generate a TBox exponential in size, but it has been proved (Nebel, 1990) that it works in polynomial time under reasonable restrictions. The following interchangeably refers either to reasoning with respect to a TBox or to reasoning involving expanded concepts with an empty TBox. In particular, while devising the subsumption calculus for the logics considered here, it is always assumed that all defined concepts have been expanded.

## 3. Towards a Temporal Description Logics

Schmiedel (1990) proposed to extend Description Logics with an interval–based temporal logic. The temporal variant of the Description Logic is equipped with a model-theoretic semantics. The underlying Description Logic is $\mathcal{FLENR}^-$ (Donini et al., 1995): it differs from $\mathcal{ALCF}$ in that it does not contain the $\top$ and $\bot$ concepts, it does not have neither negation nor disjunction, and it has cardinality restrictions and conjunction over roles. The new temporal term-forming operators are the temporal qualifier at, the existential and universal temporal quantifiers sometime and alltime. The qualifier operator specifies the time at which a concept holds. The temporal quantifiers introduce the temporal variables constrained by means of temporal relationships based on Allen's interval algebra extended with metric constraints to deal with durations, absolute times, and granularities of intervals. To give an example of this temporal Description Logic, the concept of Mortal can be defined by:

Mortal $\doteq$ LivingBeing $\sqcap$ (sometime(x) (after x NOW) (at x ($\neg$LivingBeing)))

with the meaning of a LivingBeing at the reference interval NOW, who will not be alive at an interval x sometime after the reference interval NOW. Schmiedel does not propose any algorithm for computing subsumption, but gives some preliminary hints. Actually, Schmiedel's logic is argued to be undecidable (Bettini, 1997), sacrificing the main benefit of Description Logics: the possibility of having decidable inference techniques.

Schmiedel's temporal Description Logic, when closed under complementation, contains as a proper fragment the temporal logic $\mathcal{HS}$ proposed by Halpern and Shoham (1991). The logic $\mathcal{HS}$ is a propositional modal logic which extends propositional logic with modal formulæ of the kind $\langle R \rangle.\phi$ and $[R].\phi$ – where $R$ is a basic Allen's temporal relation and $\langle \rangle$ and [] are the possibility and necessity modal operators. For example, the modal formula LivingBeing ∧ ⟨after⟩.¬LivingBeing corresponds to the abovementioned Mortal concept. Unfortunately, the $\mathcal{HS}$ logic is shown to be undecidable, at least for most interesting classes of temporal structures: *"One gets decidability only in very restricted cases, such as when the set of temporal models considered is a finite collection of structures, each consisting of a finite set of natural numbers."* (Halpern & Shoham, 1991)

Weida and Litman (1992, 1994) propose T-REX, a loose hybrid integration between Description Logics and constraint networks. Plans are defined as collections of steps together with temporal constraints between their duration. Each step is associated with an action type, represented by a generic concept in K-REP – a non-temporal Description Logic. Thus a plan is seen as a *plan network*, a temporal constraint network whose nodes, corresponding to time intervals, are labeled with action types and are associated with the steps of the plan itself. As an example of plan in T-REX they show the plan of preparing spaghetti marinara:

```
(defplan Assemble-Spaghetti-Marinara
     ((step1  Boil-Spaghetti)
      (step2  Make-Marinara)
      (step3  Put-Together-SM))
     ((step1  (before meets)  step3)
      (step2  (before meets)  step3)))
```

This is a plan composed by three actions, i.e., boiling spaghetti, preparing marinara sauce, and assembling all things at the end. Temporal constraints between the steps establish the temporal order in doing the corresponding actions. A structural plan subsumption algorithm is defined, characterized in terms of graph matching, and based on two separate notions of subsumption: pure terminological subsumption between action types labeling the nodes, and pure temporal subsumption between interval relationships labeling the arcs. The plan library is used to guide plan recognition (Weida, 1996) in a way similar to that proposed by Kautz (1991). Even if this work has strong motivations, no formal semantics is provided for the language and the reasoning problems.

Starting from the assumption that an action has a duration in time, our proposal considers an interval-based modal temporal logic – in the spirit of Halpern and Shoham (1991) – and reduces the expressivity of (Schmiedel, 1990) in the direction of (Weida & Litman, 1992). While Schmiedel's work lacks computational machinery, and Halpern and Shoham's logic is undecidable, here an expressive decidable logic is obtained, providing sound and complete reasoning algorithms. Differently from T-REX which uses two different languages to represent actions and plans – a non temporal Description Logic for describing actions and a second language to compose plans by adding temporal information – here an extension of a Description Logic is chosen in which time operators are available directly as term constructors. This view implies an integration of a temporal domain in the semantic structure where terms themselves are interpreted, giving the formal way both for a well-founded notion of subsumption and for proving soundness and completeness of the corresponding procedure. As an example of the formalism, the plan for preparing spaghetti marinara introduced above is represented as follows:

$$\texttt{Assemble-Spaghetti-Marinara} \doteq \Diamond(y\ z\ w)\ (y\ (\textsf{before}, \textsf{meets})\ w)(z\ (\textsf{before}, \textsf{meets})\ w).$$
$$(\texttt{Boil-Spaghetti}@y\ \sqcap$$
$$\texttt{Make-Marinara}@z\ \sqcap$$
$$\texttt{Put-Together-SM}@w)$$

Moreover, it is possible to build *temporal structured* actions – as opposed to the atomic actions proposed in T-REX – describing how the world state changes because of the occurrence of an action: in fact, our language allows for feature representation in order to relate actions to states of the world (see Section 5.2). This kind of expressivity is not captured by T-REX, since it uses a non-temporal Description Logic to represent actions. The main application of T-REX is plan recognition; according to the ideas of Kautz (1991) a Closed World Assumption (CWA) (Weida, 1996) is made, assuming that the plan library is complete and an observed plan will be fully accounted for by a single plan. CWA is not relied on here, following the Open World Semantics characterizing Description Logics. Weaker, but monotonic, deductions are allowed in the plan recognition process. However, their procedures for recognizing a *necessary*, *optional* or *impossible* individual plan with respect to a plan description is still applicable, if the plan library is given a closed world semantics.

## 4. The Feature Temporal Language $\mathcal{TL}$-$\mathcal{F}$

The feature temporal language $\mathcal{TL}$-$\mathcal{F}$ is the basic logic considered here. This language is composed of the temporal Logic $\mathcal{TL}$ – able to express interval temporal networks – and the non-temporal Feature Description Logic $\mathcal{F}$. Note that, each logic of the family of Temporal Description Logics introduced in this paper is identified by a composed string in which the first part refers to the temporal part of the language while the other one refers to the non-temporal part.

### 4.1 Syntax

Basic types of the language are *concepts*, *individuals*, *temporal variables* and *intervals*. Concepts can describe entities of the world, states and events. Temporal variables denote intervals bound by temporal constraints, by means of which abstract temporal patterns in the form of constraint networks are expressed. Concepts (resp. individuals) can be specified to hold at a certain temporal variable (resp. interval). In this way, *action types* (resp. *individual actions*) can be represented in a uniform way by temporally related concepts (resp. individuals).

For the basic temporal interval relations the Allen notation (Allen, 1991) (Figure 3) is used: before (b), meets (m), during (d), overlaps (o), starts (s), finishes (f), equal (=), after (a), met-by (mi), contains (di), overlapped-by (oi), started-by (si), finished-by (fi). *Concept expressions* (denoted by $C, D$) are syntactically built out of *atomic concepts* (denoted by $A$), *atomic features* (denoted by $f$), *atomic parametric features* (denoted by $\star g$) and *temporal variables* (denoted by $X, Y$). Temporal concepts $(C, D)$ are distinguished from non-temporal concepts $(E, F)$, following the syntax rules of Figure 4. Names for atomic features and atomic parametric features are from the same alphabet of symbols; the $\star$ symbol is not intended as operator, but only as differentiating the two syntactic types.

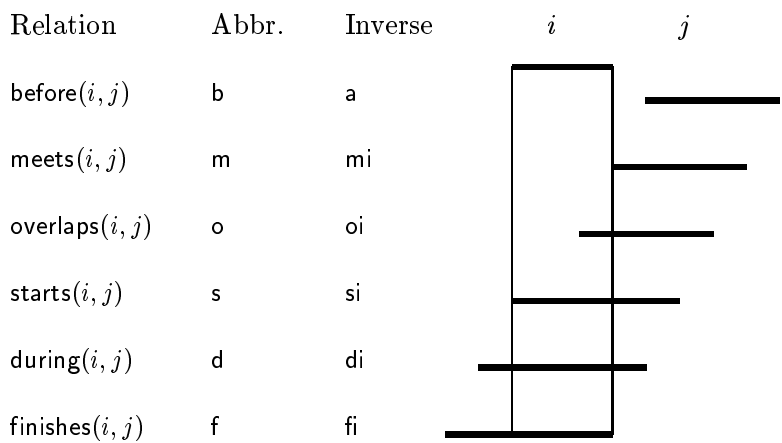

Figure 3: The Allen's interval relationships.

Temporal variables are introduced by the temporal existential quantifier "$\diamond$" – excluding the special temporal variable $\sharp$, usually called NOW, and intended as the reference interval. Variables appearing in temporal constraints ($\overline{Tc}$) must be declared within the same temporal quantifier, with the exception of the special variable $\sharp$. Temporal variables appearing in the right hand side of an "@" operator are called *bindable*. Concepts must not include *unbound* (a.k.a. *free*) bindable variables. Informally, a bindable variable is said to be *bound* in a concept if it is declared at the nearest temporal quantifier in the body of which it occurs; this avoid the usual formal inductive definition of a bound variable. Moreover, in chained constructs of the form $((C[Y_1]@X_1)[Y_2]@X_2\ldots)$ non bindable variables – i.e., the ones on the left hand side of an "@" operator – cannot appear more than once. Note that, since Description Logics are a fragment of FOL with one free variable, the above mentioned restrictions force the temporal side of the language to have only one free temporal variable, i.e., the reference time $\sharp$.

As usual, terminological axioms for building simple acyclic $\mathcal{TL}$-$\mathcal{F}$ TBoxes are allowed. While using in a concept expression a name referring to a defined concept, it is possible to use the substitutive qualifier construct, to impose a coreference with a variable appearing in the definition associated to the defined concept. The statement $C[Y]@X$ constrains the variable $Y$, which should appear in the definition of the defined concept $C$, to corefer with $X$ (see Section 5.2 for an example). A drawback in the use of this operator is the requirement to know the internal syntactical form of the defined concept, namely, the names of its temporal variables.

Let $\mathcal{O}$ and $\mathcal{OT}$ be two alphabets of symbols denoting *individuals* and *temporal intervals*, respectively. An assertion – i.e., a predication on temporally qualified individual entities – is a statement of one of the forms $C(i,a)$, $p(i,a,b)$, $\star g(a,b)$, $R(i,j)$, where $C$ is a concept, $p$ is a feature, $\star g$ is a parametric feature, $R$ is a temporal relation, $a$ and $b$ denote individuals in $\mathcal{O}$, $i$ and $j$ denote temporal intervals in $\mathcal{OT}$.

| | | | | |
|---|---|---|---|---|
| $\mathcal{TL}$ | $C, D$ | $\rightarrow$ | $E \mid$ | (non-temporal concept) |
| | | | $C \sqcap D \mid$ | (conjunction) |
| | | | $C@X \mid$ | (qualifier) |
| | | | $C[Y]@X \mid$ | (substitutive qualifier) |
| | | | $\Diamond (\overline{X})\, \overline{\mathit{Tc}}.C$ | (existential quantifier) |
| | $\mathit{Tc}$ | $\rightarrow$ | $(X\ (R)\ Y) \mid$ | (temporal constraint) |
| | | | $(X\ (R)\ \sharp) \mid$ | |
| | | | $(\sharp\ (R)\ Y)$ | |
| | $\overline{\mathit{Tc}}$ | $\rightarrow$ | $\mathit{Tc} \mid \mathit{Tc}\ \overline{\mathit{Tc}}$ | |
| | $R, S$ | $\rightarrow$ | $R\ ,\ S \mid$ | (disjunction) |
| | | | $\mathsf{s} \mid \mathsf{mi} \mid \mathsf{f} \mid \ldots$ | (Allen's relations) |
| | $X, Y$ | $\rightarrow$ | $\mathsf{x} \mid \mathsf{y} \mid \mathsf{z} \mid \ldots$ | (temporal variables) |
| | $\overline{X}$ | $\rightarrow$ | $X \mid X\ \overline{X}$ | |
| $\mathcal{F}$ | $E, F$ | $\rightarrow$ | $A \mid$ | (atomic concept) |
| | | | $\top \mid$ | (top) |
| | | | $E \sqcap F \mid$ | (conjunction) |
| | | | $p \downarrow q \mid$ | (agreement) |
| | | | $p : E$ | (selection) |
| | $p, q$ | $\rightarrow$ | $f \mid$ | (atomic feature) |
| | | | $\star g \mid$ | (atomic parametric feature) |
| | | | $p \circ q$ | (path) |

Figure 4: Syntax rules for the interval Description Logic $\mathcal{TL}$-$\mathcal{F}$

### 4.1.1 A CLARIFYING EXAMPLE

Let us now informally see the intended *meaning* of the terms of the language $\mathcal{TL}$-$\mathcal{F}$ (for the formal details see Section 4.2). Concept expressions are interpreted over pairs of *temporal intervals* and *individuals* $\langle i, a \rangle$, meaning that the individual $a$ is in the extension of the concept at the interval $i$. If a concept is intended to describe an action, then its interpretation can be seen as the set of individual actions of that type occurring at some interval.

Within a concept expression, the special "$\sharp$" variable denotes the current interval of evaluation; in the case of actions, it is thought that it refers to the temporal interval at which the action itself *occurs*. The temporal existential quantifier introduces interval variables, related to each other and possibly to the $\sharp$ variable in a way defined by the set of *temporal constraints*. To evaluate a concept at an interval $X$, different from the current one, it is necessary to temporally qualify it at $X$ – written $C@X$; in this way, every occurrence of $\sharp$ embedded within the concept expression $C$ is interpreted as the $X$ variable[2]. The informal meaning of a concept with a temporal existential quantification can be understood with the following examples in the action domain.

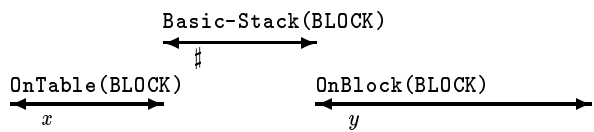

Figure 5: Temporal dependencies in the definition of the `Basic-Stack` action.

$$\texttt{Basic-Stack} \doteq \Diamond(x\ y)\ (x\ \textsf{m}\ \sharp)(\sharp\ \textsf{m}\ y).\ \bigl((\star\texttt{BLOCK} : \texttt{OnTable})@x \sqcap (\star\texttt{BLOCK} : \texttt{OnBlock})@y\bigr)$$

Figure 5 shows the temporal dependencies of the intervals in which the concept `Basic-Stack` holds. `Basic-Stack` denotes, according to the definition (a terminological axiom), any action occurring at some interval involving a $\star$BLOCK that was once `OnTable` and then `OnBlock`. The $\sharp$ interval could be understood as the occurring time of the action type being defined: referring to it within the definition is an explicit way to temporally relate states and actions occurring in the world with respect to the occurrence of the action itself. The temporal constraints $(x\ \textsf{m}\ \sharp)$ and $(\sharp\ \textsf{m}\ y)$ state that the interval denoted by $x$ should meet the interval denoted by $\sharp$ – the occurrence interval of the action type `Basic-Stack` – and that $\sharp$ should meet $y$. The parametric feature $\star$BLOCK plays the role of *formal* parameter of the action, mapping any individual action of type `Basic-Stack` to the block to be stacked, independently from time. Please note that, whereas the existence and identity of the $\star$BLOCK of the action is time invariant, it can be qualified differently in different intervals of time, e.g., the $\star$BLOCK is necessarily `OnTable` only during the interval denoted by $x$.

Let us comment now on the introduction of explicit temporal variables. The absence of explicit temporal variables would weaken the temporal structure of a concept since arbitrary relationships between more than two intervals could not be represented anymore. For example, having only implicit intervals it is not possible to describe the situation in which two concept expressions, say $C$ and $D$, hold at two meeting intervals (say $x$, $y$) with the first interval starting and the second finishing the reference interval (i.e., the temporal pattern $(x\ \textsf{meets}\ y)(x\ \textsf{starts}\ \sharp)(y\ \textsf{finishes}\ \sharp)$ cannot be represented). More precisely, it is not possible to represent temporal relations between more than two intervals if they are not derivable by the temporal propagation of the constraints imposed on pairs of variables. While explicit variables go against the general thrust of Description Logics, the gained expressive power together with the observation that the variables are limited only to the temporal part of the language are the main motivations for using them. However, it is easy to drop them by limiting the temporal expressiveness as proposed by Bettini (1997) (see also Section 9).

An assertion of the type `Basic-Stack`$(i, a)$ states that the individual $a$ is an action of the type `Basic-Stack` occurred at the interval $i$. Moreover, the same assertion implies that $a$ is related to a $\star$BLOCK, say $b$, which is of type `OnTable` at some interval $j$, meeting $i$, and of type `OnBlock` at another interval $l$, met by $i$.

---

2. Since any concept is implicitly temporally qualified at the special $\sharp$ variable, it is not necessary to explicitly qualify concepts at $\sharp$.

$$
\begin{aligned}
(\mathsf{s})^{\mathcal{E}} &= \{\langle[u,v],[u_1,v_1]\rangle \in \mathcal{T}_<^\star \times \mathcal{T}_<^\star \mid u = u_1 \wedge v < v_1\} \\
(\mathsf{f})^{\mathcal{E}} &= \{\langle[u,v],[u_1,v_1]\rangle \in \mathcal{T}_<^\star \times \mathcal{T}_<^\star \mid v = v_1 \wedge u_1 < u\} \\
(\mathsf{mi})^{\mathcal{E}} &= \{\langle[u,v],[u_1,v_1]\rangle \in \mathcal{T}_<^\star \times \mathcal{T}_<^\star \mid u = v_1\} \\
&\ldots \text{ (meaning of the other Allen temporal relations)} \\
(R\ ,\ S)^{\mathcal{E}} &= R^{\mathcal{E}} \cup S^{\mathcal{E}} \\
\langle \overline{X}, \overline{Tc} \rangle^{\mathcal{E}} &= \{\mathcal{V}: \overline{X} \mapsto \mathcal{T}_<^\star \mid \forall (X\ (R)\ Y) \in \overline{Tc}.\ \langle \mathcal{V}(X), \mathcal{V}(Y) \rangle \in R^{\mathcal{E}}\}.
\end{aligned}
$$

Figure 6: The temporal interpretation function.

$$\texttt{Basic-Stack}(i,a) \implies \exists b.\ \star\texttt{BLOCK}(a,b) \wedge \exists j, l.\ (\texttt{OnTable}(j,b) \wedge \texttt{OnBlock}(l,b) \wedge \\ \mathsf{m}(j,i) \wedge \mathsf{m}(i,l))$$

An individual action is an object in the conceptual domain associated with the relevant properties – or states – of the world affected by the individual action itself via a bunch of *features*; moreover, temporal relations constrain time intervals imposing an ordering in the change of the states of the world.

### 4.2 Semantics

In this Section, a Tarski-style extensional semantics for the $\mathcal{TL}$-$\mathcal{F}$ language is given, and a formal definition of the subsumption and recognition reasoning tasks is devised.

Assume a linear, unbounded, and dense temporal structure $\mathcal{T} = (\mathcal{P}, <)$, where $\mathcal{P}$ is a set of time points and $<$ is a strict partial order on $\mathcal{P}$. In such a structure, given an interval $X$ and a temporal relation $R$, it is always possible to find an interval $Y$ such that $(X\ (R)\ Y)$. The assumption of linear time – which means that for any two points $t_1$ and $t_2$ such that $t_1 \leq t_2$ the set of points $\{t \mid t_1 \leq t \leq t_2\}$ is totally ordered – fits the intuition about the nature of time, so that the pair $[t_1, t_2]$ can be thought as the closed interval of points between $t_1$ and $t_2$. The *interval set* of a structure $\mathcal{T}$ is defined as the set $\mathcal{T}_<^\star$ of all closed intervals $[u,v] \doteq \{x \in \mathcal{P} \mid u \leq x \leq v, u \neq v\}$ in $\mathcal{T}$.

A *primitive interpretation* $\mathcal{I} \doteq \langle \mathcal{T}_<^\star, \Delta^{\mathcal{I}}, \cdot^{\mathcal{I}} \rangle$ consists of a set $\mathcal{T}_<^\star$ (the *interval set* of the selected temporal structure $\mathcal{T}$), a set $\Delta^{\mathcal{I}}$ (the *domain* of $\mathcal{I}$), and a function $\cdot^{\mathcal{I}}$ (the *primitive interpretation function* of $\mathcal{I}$) which gives a meaning to atomic concepts, features and parametric features:

$$A^{\mathcal{I}} \subseteq \mathcal{T}_<^\star \times \Delta^{\mathcal{I}}\ ; \qquad f^{\mathcal{I}}: (\mathcal{T}_<^\star \times \Delta^{\mathcal{I}}) \stackrel{partial}{\longmapsto} \Delta^{\mathcal{I}}\ ; \qquad \star g^{\mathcal{I}}: \Delta^{\mathcal{I}} \stackrel{partial}{\longmapsto} \Delta^{\mathcal{I}}$$

Atomic parametric features are interpreted as partial functions; they differ from atomic features for being independent from time.

In order to give a meaning to temporal expressions present in generic concept expressions, Figure 6 defines the *temporal interpretation function*. The *temporal interpretation function* $\cdot^{\mathcal{E}}$ depends only on the temporal structure $\mathcal{T}$. The labeled directed graph $\langle \overline{X}, \overline{Tc} \rangle$ – where $\overline{X}$ is the set of variables representing the nodes, and $\overline{Tc}$ is the set of temporal constraints representing the arcs – is called *temporal constraint network*. The interpretation

$$
\begin{aligned}
A^{\mathcal{I}}_{\mathcal{V},t,\mathcal{H}} &= \{a \in \Delta^{\mathcal{I}} \mid \langle t, a \rangle \in A^{\mathcal{I}}\} = A^{\mathcal{I}}_t \\
\top^{\mathcal{I}}_{\mathcal{V},t,\mathcal{H}} &= \Delta^{\mathcal{I}} = \top^{\mathcal{I}} \\
(C \sqcap D)^{\mathcal{I}}_{\mathcal{V},t,\mathcal{H}} &= C^{\mathcal{I}}_{\mathcal{V},t,\mathcal{H}} \cap D^{\mathcal{I}}_{\mathcal{V},t,\mathcal{H}} \\
(p \downarrow q)^{\mathcal{I}}_{\mathcal{V},t,\mathcal{H}} &= \{a \in \mathrm{dom}\, p^{\mathcal{I}}_t \cap \mathrm{dom}\, q^{\mathcal{I}}_t \mid p^{\mathcal{I}}_t(a) = q^{\mathcal{I}}_t(a)\} = (p \downarrow q)^{\mathcal{I}}_t \\
(p : C)^{\mathcal{I}}_{\mathcal{V},t,\mathcal{H}} &= \{a \in \mathrm{dom}\, p^{\mathcal{I}}_t \mid p^{\mathcal{I}}_t(a) \in C^{\mathcal{I}}_{\mathcal{V},t,\mathcal{H}}\} \\
(C@X)^{\mathcal{I}}_{\mathcal{V},t,\mathcal{H}} &= C^{\mathcal{I}}_{\mathcal{V},\mathcal{V}(X),\mathcal{H}} \\
(C[Y]@X)^{\mathcal{I}}_{\mathcal{V},t,\mathcal{H}} &= C^{\mathcal{I}}_{\mathcal{V},t,\mathcal{H} \cup \{Y \mapsto \mathcal{V}(X)\}} \\
(\Diamond(\overline{X})\, \overline{Tc}.\, C)^{\mathcal{I}}_{\mathcal{V},t,\mathcal{H}} &= \{a \in \Delta^{\mathcal{I}} \mid \exists \mathcal{W}.\, \mathcal{W} \in \langle \overline{X}, \overline{Tc} \rangle^{\mathcal{E}}_{\mathcal{H} \cup \{\sharp \mapsto t\}} \wedge a \in C^{\mathcal{I}}_{\mathcal{W},t,\emptyset}\} \\
f^{\mathcal{I}}_t &= \hat{f}_t : \Delta^{\mathcal{I}} \stackrel{partial}{\longmapsto} \Delta^{\mathcal{I}} \mid \forall a.\, (a \in \mathrm{dom}\, \hat{f}_t \leftrightarrow \langle t, a \rangle \in \mathrm{dom}\, f^{\mathcal{I}}) \wedge \\
&\qquad \hat{f}_t(a) = f^{\mathcal{I}}(t, a) \\
(p \circ q)^{\mathcal{I}}_t &= p^{\mathcal{I}}_t \circ q^{\mathcal{I}}_t \\
\star g^{\mathcal{I}}_t &= \star g^{\mathcal{I}}
\end{aligned}
$$

Figure 7: The interpretation function.

of a temporal constraint network is a set of variable assignments that satisfy the temporal constraints. A *variable assignment* is a function $\mathcal{V} : \overline{X} \mapsto \mathcal{T}^{\star}_{<}$ associating an interval value to a temporal variable. A temporal constraint network is *consistent* if it admits a non empty interpretation. The notation, $\langle \overline{X}, \overline{Tc} \rangle^{\mathcal{E}}_{\{x_1 \mapsto t_1, x_2 \mapsto t_2, \ldots\}}$, used to interpret concept expressions, denotes the subset of $\langle \overline{X}, \overline{Tc} \rangle^{\mathcal{E}}$ where the variable $x_i$ is mapped to the interval value $t_i$.

It is now possible to interpret generic concept expressions. Consider the equations introduced in Figure 7. An *interpretation function* $\cdot^{\mathcal{I}}_{\mathcal{V},t,\mathcal{H}}$, based on a variable assignment $\mathcal{V}$, an interval $t$ and a set of constraints $\mathcal{H} = \{x_1 \mapsto t_1, \ldots\}$ over the assignments of inner variables, extends the primitive interpretation function in such a way that the equations of Figure 7 are satisfied. Intuitively, the interpretation of a concept $C^{\mathcal{I}}_{\mathcal{V},t,\mathcal{H}}$ is the set of entities of the domain that are of type $C$ at the time interval $t$, with the assignment for the free temporal variables in $C$ given by $\mathcal{V}$ – see $(C@X)^{\mathcal{I}}_{\mathcal{V},t,\mathcal{H}}$ – and with the constraints for the assignment of variables in the scope of the outermost temporal quantifiers given by $\mathcal{H}$. Note that, $\mathcal{H}$ interprets the variable renaming due to the temporal substitutive qualifier – see $(C[Y]@X)^{\mathcal{I}}_{\mathcal{V},t,\mathcal{H}}$ – and it takes effect during the choice of a variable assignment, as the equation for $(\Diamond(\overline{X})\, \overline{Tc}.\, C)^{\mathcal{I}}_{\mathcal{V},t,\mathcal{H}}$ shows.

In absence of free variables in the concept expression – with the exception of $\sharp$ – for notational simplification the *natural* interpretation function $C^{\mathcal{I}}_t$, being equivalent to the interpretation function $C^{\mathcal{I}}_{\mathcal{V},t,\mathcal{H}}$ with any $\mathcal{V}$ such that $\mathcal{V}(\sharp) = t$ and $\mathcal{H} = \emptyset$ is introduced. The set of interpretations $\{C^{\mathcal{I}}_{\mathcal{V},t,\mathcal{H}}\}$ obtained by varying $\mathcal{I}, \mathcal{V}, t$ with a fixed $\mathcal{H}$ is maximal wrt set inclusion if $\mathcal{H} = \emptyset$, i.e., the set of natural interpretations includes any set of interpretations with a fixed $\mathcal{H}$. In fact, since $\mathcal{H}$ represents a constraint in the assignment of variables, the unconstrained set is the larger one. Note that, for feature interpretation only the natural one is used since it is not admitted to temporally qualify them.

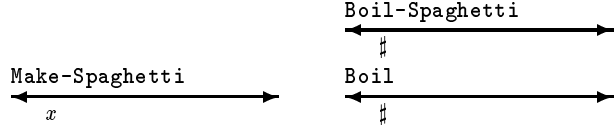

Figure 8: Temporal dependencies in the definition of the Boil-Spaghetti plan.

An interpretation $\mathcal{I}$ satisfies the terminological axiom $A \doteq C$ iff $A_t^\mathcal{I} = C_t^\mathcal{I}$, for every $t$. A concept $C$ is *subsumed* by a concept $D$ ($C \sqsubseteq D$) if $C_t^\mathcal{I} \subseteq D_t^\mathcal{I}$ for every interpretation $\mathcal{I}$ and every interval $t$. An interpretation $\mathcal{I}$ is a *model* for a concept $C$ if $C_t^\mathcal{I} \neq \emptyset$ for some $t$. If a concept has a model, then it is *satisfiable*, otherwise it is *unsatisfiable*.

Each $\mathcal{TL}\text{-}\mathcal{F}$ concept expression is always satisfiable, with the proviso that the temporal constraints introduced by the existential quantifiers are consistent. This latter condition can be easily checked during the reduction of the concept into a *normal form* when the *minimal* temporal network (see Section 11, definition 6.5) is computed.

It is interesting to note that only the relations s, f, mi are really necessary, because it is possible to express any temporal relationship between two distinct intervals using only these three relations and their transpositions si, fi, m (Halpern & Shoham, 1991). The following equivalences hold:

$\Diamond x\ (x\ \text{a}\ \sharp).\ C@x\ \equiv\ \Diamond xy\ (y\ \text{mi}\ \sharp)(x\ \text{mi}\ y).\ C@x$

$\Diamond x\ (x\ \text{d}\ \sharp).\ C@x\ \equiv\ \Diamond xy\ (y\ \text{s}\ \sharp)(x\ \text{f}\ y).\ C@x$

$\Diamond x\ (x\ \text{o}\ \sharp).\ C@x\ \equiv\ \Diamond xy\ (y\ \text{s}\ \sharp)(x\ \text{fi}\ y).\ C@x$

To assign a meaning to ABox axioms, the temporal interpretation function $\cdot^\mathcal{E}$ is extended to temporal intervals so that $i^\mathcal{E}$ is an element of $\mathcal{T}_<^\star$ for each $i \in \mathcal{OT}$. The semantics of assertions is the following: $C(i, a)$ is satisfied by an interpretation $\mathcal{I}$ iff $a^\mathcal{I} \in C_{i^\mathcal{E}}^\mathcal{I}$; $p(i, a, b)$ is satisfied by $\mathcal{I}$ iff $p_{i^\mathcal{E}}^\mathcal{I}(a^\mathcal{I}) = b^\mathcal{I}$; $\star g(a, b)$ is satisfied by $\mathcal{I}$ iff $\star g^\mathcal{I}(a^\mathcal{I}) = b^\mathcal{I}$; and $R(i, j)$ is satisfied by $\mathcal{I}$ iff $\langle i^\mathcal{E}, j^\mathcal{E} \rangle \in R^\mathcal{E}$. Given a knowledge base $\Sigma$, an individual $a$ in $\mathcal{O}$ is said to be an *instance* of a concept $C$ *at the interval* $i$ if $\Sigma \models C(i, a)$.

Now we are able to give a semantic definition for the reasoning task we already called *specific plan recognition with respect to a plan description*. This is an inference service that computes if an *individual action/plan* is an instance of an *action/plan type* at a certain interval, i.e., the task known as instance recognition in the Description Logic community. Given a knowledge base $\Sigma$, an interval $i$, an individual $a$ and a concept $C$, the *instance recognition problem* is to test whether $\Sigma \models C(i, a)$.

## 5. Action and plan representation: two examples

An action description represents how the world state may evolve in relation with the possible occurrence of the action itself. A plan is a complex action: it is described by means of temporally related world states and simpler actions. The following introduces examples of action and plan representations from two well known domains, the cooking domain (Kautz, 1991; Weida & Litman, 1992) and the block world (Allen, 1991), with the aim of showing the applicability of our framework.

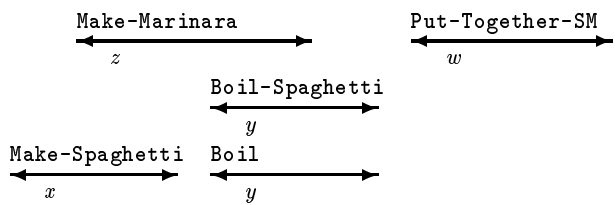

Figure 9: Temporal dependencies in the definition of Assemble-Spaghetti-Marinara.

### 5.1 The Cooking Domain

Let us introduce the plan Boil-Spaghetti:

Boil-Spaghetti $\doteq$ $\diamond x$ $(x$ b $\sharp)$. (Make-Spaghetti@$x$ ⊓ Boil)

Figure 8 shows the temporal dependencies of the intervals in which the concept Boil-Spaghetti holds. The definition employs the $\sharp$ interval to denote the occurrence time of the plan itself; in this way, it is possible to describe how different actions or states of the world concurring to the definition of the plan are related to it. This is why the variable $\sharp$ is explicitly present in the definition of Boil-Spaghetti, instead of a generic variable: the Boil action should take place at the same time of the plan itself, while Make-Spaghetti occurs before it.

The definition of a plan can be reused within the definition of other plans by exploiting the full compositionality of the language. The plan defined above Boil-Spaghetti is used in the definition of Assemble-Spaghetti-Marinara:

Assemble-Spaghetti-Marinara $\doteq$ $\diamond(y\ z\ w)$ $(y$ b $w)(z$ b $w)$.
                                        (Boil-Spaghetti@$y$ ⊓
                                         Make-Marinara@$z$ ⊓
                                         Put-Together-SM@$w$)

In this case, precise temporal relations between the nodes of two corresponding temporal constraint networks are asserted: e.g., the action Put-Together-SM takes place strictly after the Boil action (Figure 9). Observe that the occurrence interval of the plan Assemble-Spaghetti-Marinara does not appear in the Figure because it is not temporally related with any other interval.

A plan subsuming Assemble-Spaghetti-Marinara is the more general plan defined below, Prepare-Spaghetti, supposing that the action Make-Sauce subsumes Make-Marinara. This means that among all the individual actions of the type Prepare-Spaghetti there are all the individual actions of type Assemble-Spaghetti-Marinara:

Prepare-Spaghetti $\doteq$ $\diamond$ $(y\ z)$ (). (Boil-Spaghetti@$y$ ⊓ Make-Sauce@$z$)

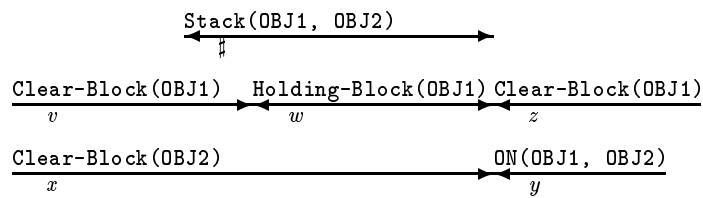

Figure 10: Temporal dependencies in the definition of the Stack action.

However, note that Boil-Spaghetti does not subsume Prepare-Spaghetti, even if it is a conjunct in the definition of the latter. This could be better explained observing how the definition of Prepare-Spaghetti plan is expanded:

Prepare-Spaghetti $\doteq$ $\diamond$ $(x\ y\ z)\ (x\ \mathsf{b}\ y)$. (Make-Spaghetti@$x$ ⊓ Boil@$y$ ⊓ Make-Sauce@$z$)

Then, the Boil action occurs at the interval $y$ – which can be different from the occurring time of Prepare-Spaghetti – as the effect of binding Boil-Spaghetti to the temporal variable $y$. On the contrary, in the definition of Boil-Spaghetti the Boil action takes place *necessarily* at the same time. Subsumption between Prepare-Spaghetti and Boil-Spaghetti fails since different temporal relations between the actions describing the two plans and the plans themselves are specified. In particular, observe that the Boil-Spaghetti plan denotes a narrower class than the plan expression

$\diamond(x\ y)\ (x\ \mathsf{b}\ y)$. (Make-Spaghetti@$x$ ⊓ Boil@$y$)

which subsumes both Prepare-Spaghetti and Boil-Spaghetti itself.

### 5.2 The Blocks World Domain

As a further example of the expressive power of the temporal language, it is now shown how to represent the Stack action in the blocks world, in a more detailed way than the previous simple Basic-Stack action used as a clarifying example. Thus a stacking action involves two blocks, which should be both clear at the beginning; the central part of the action consists of grasping one block; at the end, the blocks are one on top of another, and the bottom one is no longer clear (Figure 10).

Our representation borrows from the RAT Description Logic (Heinsohn, Kudenko, Nebel, & Profitlich, 1992) the intuition of representing action parameters by means of partial functions mapping from the action itself to the involved action parameter (see Section 9). In the language, these functions are called *parametric features*. For example, the action Stack has the parameters ⋆OBJECT1 and ⋆OBJECT2, representing in some sense the objects that are involved in the action independently from time. So, in the assertion "⋆OBJECT1$(a, block\text{-}a)$", *block-a* denotes the first object involved in the action $a$ at any interval. On the other hand, an assertion involving a (non-parametric) feature, e.g., "ON$(i, block\text{-}a, block\text{-}b)$", does not imply anything about the truth value at intervals other than $i$.

The concept expression, which defines the Stack action, makes use of temporal qualified concept expressions, including feature *selections* and *agreements*: the expression (⋆OBJECT2 : Clear-Block)@$x$ means that the second parameter of the action should be a Clear-Block at the interval denoted by $x$; while $(\star\texttt{OBJECT1} \circ \texttt{ON} \downarrow \star\texttt{OBJECT2})@y$ indicates that at the interval $y$ the object on which $\star\texttt{OBJECT1}$ is placed is $\star\texttt{OBJECT2}$. The formal definition of the action Stack is:

$$\texttt{Stack} \doteq \Diamond(x\ y\ z\ v\ w)\ (x\ \textsf{fi}\ \sharp)(y\ \textsf{mi}\ \sharp)(z\ \textsf{mi}\ \sharp)(v\ \textsf{o}\ \sharp)(w\ \textsf{f}\ \sharp)(w\ \textsf{mi}\ v).$$
$$((\star\texttt{OBJECT2}:\texttt{Clear-Block})@x \sqcap (\star\texttt{OBJECT1} \circ \texttt{ON} \downarrow \star\texttt{OBJECT2})@y \sqcap$$
$$(\star\texttt{OBJECT1}:\texttt{Clear-Block})@v \sqcap (\star\texttt{OBJECT1}:\texttt{Holding-Block})@w \sqcap$$
$$(\star\texttt{OBJECT1}:\texttt{Clear-Block})@z)$$

The above defined concept does not state which properties are the prerequisites for the stacking action or which properties must be true whenever the action succeeds. What this action intuitively states is that $\star\texttt{OBJECT1}$ will be on $\star\texttt{OBJECT2}$ in a situation where both objects are clear at the start of the action. Note that the world states described at the intervals denoted by $v, w, z$ are the result of an action of *grasping* a previously clear block:

$$\texttt{Grasp} \doteq \Diamond(x\ w\ z)\ (x\ \textsf{o}\ \sharp)(w\ \textsf{f}\ \sharp)(w\ \textsf{mi}\ x)(z\ \textsf{mi}\ \sharp).$$
$$((\star\texttt{OBJECT1}:\texttt{Clear-Block})@x \sqcap (\star\texttt{OBJECT1}:\texttt{Holding-Block})@w \sqcap$$
$$(\star\texttt{OBJECT1}:\texttt{Clear-Block})@z)$$

The Stack action can be redefined by making use of the Grasp action:

$$\texttt{Stack} \doteq \Diamond(x\ y\ u\ v)\ (x\ \textsf{fi}\ \sharp)(y\ \textsf{mi}\ \sharp)(u\ \textsf{f}\ \sharp)(v\ \textsf{o}\ \sharp).$$
$$((\star\texttt{OBJECT2}:\texttt{Clear-Block})@x \sqcap (\star\texttt{OBJECT1} \circ \texttt{ON} \downarrow \star\texttt{OBJECT2})@y \sqcap$$
$$(\texttt{Grasp}[x]@v)@u)$$

The temporal substitutive qualifier $(\texttt{Grasp}[x]@v)$ *renames* within the defined Grasp action the variable $x$ to $v$ and it is a way of making coreference between two temporal variables, while the temporal constraints peculiar to the renamed variable $x$ are inherited by the substituting interval $v$. Furthermore, the effect of temporally qualifying the grasping action at $u$ is that the $\sharp$ variable associated to the grasping action – referring to the occurrence time of the action itself – is bound to the interval denoted by $u$. Because of this binding on the occurrence time of the grasping action, the $\sharp$ variable in the grasping action and the $\sharp$ variable in the stacking action denote different time intervals, so that the grasping action occurs at an interval finishing the occurrence time of the stacking action.

Now it is shown how from a series of outside observations action recognition can be performed – i.e., the task called *specific plan recognition with respect to a plan description*. The following ABox describes a situation in which blocks can be *clear*, *grasped* and/or *on* each other, and in which a generic individual action $a$ is taking place at time interval $i_a$ having the blocks *block-a* and *block-b* as its parameters:

$\star\texttt{OBJECT1}(a, \textit{block-a}),\ \star\texttt{OBJECT2}(a, \textit{block-b}),$

$\textsf{o}(i_1, i_a),\ \texttt{Clear-Block}(i_1, \textit{block-a}),\ \textsf{fi}(i_2, i_a),\ \texttt{Clear-Block}(i_2, \textit{block-b}),$

$\textsf{mi}(i_3, i_1),\ \textsf{f}(i_3, i_a),\ \texttt{Holding-Block}(i_3, \textit{block-a}),$

$\textsf{mi}(i_4, i_a),\ \texttt{Clear-Block}(i_4, \textit{block-a}),\ \textsf{mi}(i_5, i_a),\ \texttt{ON}(i_5, \textit{block-a}, \textit{block-b})$

The system deduces that, in the context of a knowledge base $\Sigma$ composed by the above ABox and the definition of the Stack concept in the TBox, the individual action $a$ is of type Stack at the time interval $i_a$, i.e., $\Sigma \models \texttt{Stack}(i_a, a)$.

$$
\begin{aligned}
C@X \sqcap D@X &\to (C \sqcap D)@X \\
(C@X_1)@X_2 &\to C@X_1 \\
(C@X_1 \sqcap D)@X_2 &\to C@X_1 \sqcap D@X_2 \\
C \sqcap \Diamond(\overline{X})\,\overline{\mathit{Tc}}.\,D &\to \Diamond(\overline{X})\,\overline{\mathit{Tc}}.\,(C \sqcap D) \\
&\quad \text{if } C \text{ doesn't contain free variables} \\
\big[\big(\Diamond(\overline{Y})\,\overline{\mathit{Tc}}_2.\,D\big)[Y_1]@X_1\ldots[Y_p]@X_q\big]@X &\to \Diamond(\overline{X}\uplus_{[Y_1/X_1]\ldots[Y_p/X_q]}\overline{Y})\overline{\mathit{Tc}}_1\cup\overline{\mathit{Tc}}_{2+_{[\sharp/X]}}.(C \sqcap D_+@X) \\
\Diamond(\overline{X})\overline{\mathit{Tc}}_1.\big(C\sqcap \phantom{x}\big) & \\
&\quad \text{if } D \text{ doesn't contain existential temporal quantifiers} \\
p : (q : C) &\to (p \circ q) : C \\
p : (C \sqcap D) &\to p : C \sqcap p : D \\
p : (q_1 \downarrow q_2) &\to p \circ q_1 \downarrow p \circ q_2
\end{aligned}
$$

*Prescriptions:* $\overline{X}\uplus_{[Y_1/X_1]\ldots[Y_p/X_q]}\overline{Y}$ returns the union of the two sets of variables $\overline{X}$ and $\overline{Y}$, where each occurrence of $Y_1,\ldots,Y_p$ is substituted by $X_1,\ldots,X_q$, respectively, while all the other elements of $\overline{Y}$ occurring in $\overline{X}$ are renamed with fresh new identifiers. $Z_+$ is intended to be the expression $Z$ where the same substitution or renaming has taken place. The condition on the last rule forces application to start from the last nested existential temporal qualified concept.

Figure 11: Rewrite rules to transform an arbitrary concept into an existential concept.

## 6. The Calculus for $\mathcal{TL}\text{-}\mathcal{F}$

This Section presents a calculus for deciding subsumption between temporal concepts in the Description Logic $\mathcal{TL}\text{-}\mathcal{F}$. The calculus is based on the idea of separating the inference on the temporal part from the inference on the Description Logic part. This is achieved by first looking for a *normal form* of concepts. Concept subsumption in the temporal language is then reduced to concept subsumption between non-temporal concepts and to subsumption between temporal constraint networks.

### 6.1 Normal Form

Every $\mathcal{TL}\text{-}\mathcal{F}$ concept expression can be reduced to an equivalent *existential concept* of the form: $\Diamond(\overline{X})\,\overline{\mathit{Tc}}.\,(Q_0 \sqcap Q_1@X_1 \sqcap \ldots \sqcap Q_n@X_n)$, where each $Q$ is a non-temporal concept, i.e., it is an element of the language $\mathcal{F}$. A concept in existential form can be seen as a *conceptual temporal constraint network*, i.e., a labeled directed graph $\langle \overline{X}, \overline{\mathit{Tc}}, \overline{Q@X} \rangle$ where arcs are labeled with a set of arbitrary temporal relationships – representing their disjunction – nodes are labeled with non-temporal concepts and, for each node $X$, the temporal relation $(X = X)$ is implicitly true. Moreover, since the normalized concepts do not contain free variables or substitutive qualifiers, in the following the natural interpretation function (see Section 4.2) is used.

**Proposition 6.1 (Equivalence of EF)** *Every concept $C$ can be reduced in linear time into an equivalent existential concept (ef $C$), by exhaustively applying the set of rewrite rules of Figure 11.*

```
Procedure Covering(⟨X̄, T̄c⟩, y):
   mid ⟵ ∅;
   result ⟵ ∅;
   Z̄ = {z ∈ X̄ | (z (=,...) y) ∈ T̄c};
   ∀s ∈ ℘(Z̄) do
      if |s| ≥ 2 and the graph ⟨X̄, T̄c*⟩ obtained by deleting the "=" temporal relation between
      the node y and each of the nodes in s is inconsistent
         then mid ⟵ mid ∪ {s};
   ∀s ∈ mid do
      if ¬∃t ∈ mid. t ⊆ s
         then result ⟵ result ∪ {s};
   return result.
```

Figure 12: Procedure which computes a covering.

Note that (ef C) makes explicit all the possible chains of features by reducing each non-temporal concept Q to a conjunction of atomic concepts, feature selections restricted to atomic concepts and feature agreements – i.e., each Q is a feature term expression (Smolka, 1992).

The normalization proceeds by discovering all the possibles interactions between nodes with the intention of making explicit all the implicit information. A crucial temporal interaction occurs when a node is always coincident with a set of nodes in every possible interpretation of the temporal network.

**Definition 6.2 (Covering)** *Given a temporal constraint network $\langle \overline{X}, \overline{Tc} \rangle$, let $y \in \overline{X}$ and $\overline{Z} = \{z_1, z_2, \ldots, z_p\} \subseteq \overline{X}$, with $p \geq 1$, and $y \notin \overline{Z}$. $\overline{Z}$ is a Covering for $y$ if $\forall \mathcal{V} \in \langle \overline{X}, \overline{Tc} \rangle^{\mathcal{E}}$, $\mathcal{V}(y) \in \{\mathcal{V}(z_1), \mathcal{V}(z_2), \ldots, \mathcal{V}(z_p)\}$ and for each $\overline{W} \subset \overline{Z}$, $\overline{W}$ is not a covering for $y$. If $\overline{Z} = \emptyset$, then $y$ is called uncovered, otherwise $y$ is said covered by $\overline{Z}$.*

**Proposition 6.3 (Covering procedure)** *Given a temporal constraint network $\langle \overline{X}, \overline{Tc} \rangle$ in minimal form (see, e.g., (van Beek & Manchak, 1996)) and a node $y \in \overline{X}$ then the procedure described in Figure 12 returns all the possible coverings for $y$ with size $\geq 2$.*

The idea behind the covering is that whenever a set of nodes $\{z_1, z_2, \ldots, z_p\}$ is a covering for $y$ the disjunctive concept expression $(Q_{z_1} \sqcup \ldots \sqcup Q_{z_p})$ should be conjunctively added to the concept expression $Q_y$. Actually, since in $\mathcal{TL}$-$\mathcal{F}$ concept disjunction is not allowed it will be sufficient to add to the node $y$ the *Least Commom Subsumer* (LCS) of $(Q_{z_1} \sqcup \ldots \sqcup Q_{z_p})$ as defined below.

**Definition 6.4 (LCS)** *Let $Q_1, \ldots, Q_n, Q, C$ be $\mathcal{F}$ concept expressions. Then, the concept $Q = \mathrm{LCS}\{Q_1, \ldots, Q_n\}$ is such that: $Q_1 \sqsubseteq Q \wedge \ldots \wedge Q_n \sqsubseteq Q$ and there is no $C$ such that $Q_1 \sqsubseteq C \wedge \ldots \wedge Q_n \sqsubseteq C \wedge C \sqsubset Q$.*

Given a concept in existential form, the temporal completion of the constraint network is computed as described below.

**Definition 6.5 (Completed existential form)** *The temporal completion of a concept in existential form – the Completed Existential Form, CEF – is obtained by sequentially applying the following steps:*

- (closure) The transitive closure of the Allen temporal relations in the conceptual temporal constraint network is computed, obtaining a minimal temporal network (see, e.g., (van Beek & Manchak, 1996)).

- (= collapsing) For each equality temporal constraint, collapse the equal nodes by applying the following rewrite rule:
$$\Diamond(\overline{X})\ \overline{Tc}\ (x_i = x_j).\ Q \to \begin{cases} \Diamond(\overline{X} \setminus \{x_j\})\ \overline{Tc}_{[x_j/x_i]} \cdot Q_{[x_j/x_i]} & \text{if } x_i \neq x_j \text{ and } x_j \neq \natural. \\ \Diamond(\overline{X} \setminus \{x_i\})\ \overline{Tc}_{[x_i/\natural]} \cdot Q_{[x_i/\natural]} & \text{if } x_i \neq x_j \text{ and } x_j = \natural. \end{cases}$$
Then apply exhaustively the first rule of Figure 11.

- (covering) For each $y \in \overline{X}$ let compute the covering = $\{\overline{Z}_1, \ldots, \overline{Z}_n\}$ following the procedure showed by proposition 6.3. Whenever the covering is not empty, translate $Q_y$ applying the following rewrite rule: $Q_y \to Q_y \sqcap_{i=1\ldots n} \text{LCS}\{Q_{i_1}, \ldots, Q_{i_m}\}$ where $\overline{Z}_i = \{z_{i_1}, \ldots, z_{i_m}\}$, and $Q_{i_j}@z_{i_j} \in \langle \overline{X}, \overline{Tc}, \overline{Q@X} \rangle$.

- (parameter introduction) New information is added to each node because of the presence of parameters, as the following rules show. The $\leadsto$ symbol is intended so that, each time the concept expression in the left hand side appears in some node of the temporal constraint network, possibly conjoined with other concepts, then the right hand side represents the concept expression that must be conjunctively added to all the other nodes; square brackets point out optional parts; the letters $f$ ($\star f$) and $g$ ($\star g$), possibly with subscripts, denote atomic (parametric) features while $p$ and $q$ stand for generic features.

$$\begin{aligned}
\star g_1 \circ \ldots \circ \star g_n\ [\circ\ f\ [\circ\ p]] : C &\leadsto \star g_1 \circ \ldots \circ \star g_n : \top. \\
\star g_1 \circ \ldots \circ \star g_n\ [\circ\ f\ [\circ\ p]] \downarrow g\ [\circ\ q] &\leadsto \star g_1 \circ \ldots \circ \star g_n : \top. \\
\star g_1 \circ \ldots \circ \star g_n \downarrow \star f_1 \circ \ldots \circ \star f_m &\leadsto \star g_1 \circ \ldots \circ \star g_n \downarrow \star f_1 \circ \ldots \circ \star f_m. \\
\star g_1 \circ \ldots \circ \star g_n \circ g\ [\circ\ p] \downarrow \star f_1 \circ \ldots \circ \star f_m\ [\circ\ f\ [\circ\ q]] &\leadsto \star g_1 \circ \ldots \circ \star g_n : \top\ \sqcap \\
& \quad \star f_1 \circ \ldots \circ \star f_m : \top.
\end{aligned}$$

**Proposition 6.6 (Equivalence of CEF)** *Every concept in existential form can be reduced into an equivalent completed existential concept.*

Both the *covering* and the *parameter introduction* steps can be computed independently after the *=-collapsing* step and then conjoining the resulting concept expressions. Observe that, to obtain a completed existential concept, the steps of the normalization procedure require linear time with the exception of the computation of the transitive closure of the temporal relations, and the covering step. Both these steps involve NP-complete temporal constraint problems (van Beek & Cohen, 1990). However, it is possible to devise reasonable subsets of Allen's algebra for which the problem is polynomial (Renz & Nebel, 1997). The most relevant properties of a concept in CEF is that all the admissible interval temporal relations are explicit and the concept expression in each node is no more refinable without changing the overall concept meaning; this is stated by the following proposition.

**Proposition 6.7 (Node independence of CEF)** *Let $\langle \overline{X}, \overline{Tc}, \overline{Q@X} \rangle$ be a conceptual temporal constraint network in its completed form (CEF); then, for all $Q \in \overline{Q}$ and for all* $\mathcal{F}$ concept expressions $C$ such that $C \not\sqsupseteq Q$, there exists an interpretation $\mathcal{I}$ such that $\langle \overline{X}, \overline{Tc}, \overline{(Q \sqcap C)@X} \rangle^{\mathcal{I}}_t \neq \langle \overline{X}, \overline{Tc}, \overline{Q@X} \rangle^{\mathcal{I}}_t$, for some interval $t$.

*Proof.* The proposition states that the information in each node of the CEF is independent from the information in the other nodes. In fact, $\langle \overline{X}, \overline{Tc}, \overline{(Q \sqcap C)@X} \rangle^{\mathcal{I}}_t = \langle \overline{X}, \overline{Tc}, \overline{Q@X} \rangle^{\mathcal{I}}_t$ if the concept expression in one node implies new information in some other node. Two cases can be distinguished.

i) *Covered Nodes.* Both the *(= collapsing)* rule and the *(covering)* rule provide to restrict a covered node with the most specific $\mathcal{F}$ concept expression. Indeed, the *(= collapsing)* rule provides collapsing two contemporary nodes conjoining the concept expressions of each of them. On the other hand, the *(covering)* rule adds to the covered node the most specific $\mathcal{F}$ concept expression that subsumes the disjunctive concept expression that is implicitly true at the covered node. Note that, thanks to the *(Closure)* rule, all the possible equal temporal relations are made explicit. So these two normalization rules cover all the possible cases of temporal interactions between nodes.

ii) *No coincident nodes.* Every time-invariant information should spread over all the nodes. Both parametric features and the $\top$ concept have a time-invariant semantics: the only time-invariant concept expressions are $\top, \star g_1 \circ \ldots \circ \star g_n : \top, \star g_1 \circ \ldots \circ \star g_n \downarrow \star f_1 \circ \ldots \circ \star f_m$, with $n, m \geq 1$, or an arbitrary conjunction of these terms. The *(parameter introduction)* rule captures all the possible syntactical cases of completion concerning time-invariant concept expressions. By induction on the syntax, it can be proven that adding to a node any other concept expression changes the overall interpretation. $\square$

The last normalization procedure eliminates nodes with redundant information. This final normalization step ends up with the concept in the *essential graph form*, that will be the *normal form* used for checking concept subsumption.

**Definition 6.8 (Essential graph)** *The subgraph of the CEF of a conceptual temporal constraint network $T = \langle \overline{X}, \overline{Tc}, \overline{Q@X} \rangle$ obtained by deleting the nodes labeled only with time-invariant concept expressions – with the exception of the $\sharp$ node – is called* essential graph *of $T$: (ess $T$).*

**Proposition 6.9 (Equivalence of essential graph)** *Every concept in completed existential form can be reduced in linear time into an equivalent essential graph form.*

**Theorem 6.10 (Equivalence of normal form)** *Every concept expression can be reduced into an equivalent essential graph form. If a polynomial fragment of Allen's algebra is adopted, the reduction takes polynomial time.*

As an example, the normal form is shown – i.e., the essential graph – of the previously introduced `Stack` action (see Section 5.2):

```
Stack ≐ ◇(x y v w z)(x fi ♯)(y mi ♯)(z mi ♯)(w f ♯)(v o ♯)(y mi x)(z mi x)(w f x)
              (v (o,d,s) x)(z (=,s,si) y)(w m y)(v b y)(w m z)(v b z)(w mi v).
         ((⋆OBJECT2 : Clear-Block ⊓ ⋆OBJECT1 : ⊤)@x ⊓
          (⋆OBJECT1∘ON ↓ ⋆OBJECT2)@y ⊓
          (⋆OBJECT1 : Clear-Block ⊓ ⋆OBJECT2 : ⊤)@v ⊓
          (⋆OBJECT1 : Hold-Block ⊓ ⋆OBJECT2 : ⊤)@w ⊓
          (⋆OBJECT1 : Clear-Block ⊓ ⋆OBJECT2 : ⊤)@z)
```

In this example, the essential graph is also the CEF of `Stack` since there are no redundant nodes.

### 6.2 Computing Subsumption

A concept subsumes another one just in case every possible instance of the second is also an instance of the first, for every time interval. Thanks to the normal form, concept subsumption in the temporal language is reduced to concept subsumption between non-temporal concepts and to subsumption between temporal constraint networks. A similar general procedure was first presented in (Weida & Litman, 1992), where the language for non-temporal concepts is less expressive – it does not include features or parametric features.

To compute subsumption between non-temporal concepts – which may possibly include LCS concepts – we refer to (Cohen, Borgida, & Hirsh, 1992). In the following, we will write "$\sqsupseteq_\mathcal{F}$" for subsumption between non-temporal $\mathcal{F}$ concepts taking into account LCS concepts.

**Definition 6.11 (Variable mapping)** *A variable mapping $\mathcal{M}$ is a total function $\mathcal{M} : \overline{X}_1 \mapsto \overline{X}_2$ such that $\mathcal{M}(\sharp) = \sharp$. We write $\mathcal{M}(\overline{X})$ to intend $\{\mathcal{M}(X) \mid X \in \overline{X}\}$, and $\mathcal{M}(\overline{Tc})$ to intend $\{(\mathcal{M}(X) \ (R) \ \mathcal{M}(Y)) \mid (X \ (R) \ Y) \in \overline{Tc}\}$.*

**Definition 6.12 (Temporal constraint subsumption)** *A temporal constraint $(X_1(R_1)Y_1)$ is said to* subsume *a temporal constraint $(X_2(R_2)Y_2)$ under a generic variable mapping $\mathcal{M}$, written $(X_1(R_1)Y_1) \sqsupseteq_\mathcal{M} (X_2 \ (R_2) \ Y_2)$, if $\mathcal{M}(X_1) = X_2$, $\mathcal{M}(Y_1) = Y_2$ and $(R_1)^\mathcal{E} \supseteq (R_2)^\mathcal{E}$ for every temporal interpretation $\mathcal{E}$.*

**Proposition 6.13 (TC subsumption algorithm)** *$(X_1(R_1)Y_1) \sqsupseteq_\mathcal{M} (X_2(R_2)Y_2)$ if and only if $\mathcal{M}(X_1) = X_2$, $\mathcal{M}(Y_1) = Y_2$ and the disjuncts in $R_1$ are a superset of the disjuncts in $R_2$.*

*Proof.* Follows from the observation that the 13 temporal relations are mutually disjoint and their union covers the whole interval pairs space. $\square$

**Definition 6.14 (Temporal constraint network subsumption)** *A temporal constraint network $\langle \overline{X}_1, \overline{Tc}_1 \rangle$ subsumes a temporal constraint network $\langle \overline{X}_2, \overline{Tc}_2 \rangle$ under a variable mapping $\mathcal{M} : \overline{X}_1 \mapsto \overline{X}_2$, written $\langle \overline{X}_1, \overline{Tc}_1 \rangle \sqsupseteq_\mathcal{M} \langle \overline{X}_2, \overline{Tc}_2 \rangle$, if $\langle \mathcal{M}(\overline{X}_1), \mathcal{M}(\overline{Tc}_1) \rangle^\mathcal{E} \supseteq \langle \overline{X}_2, \overline{Tc}_2 \rangle^\mathcal{E}$ for every temporal interpretation $\mathcal{E}$.*

**Proposition 6.15 (TCN subsumption algorithm)** *$\langle \overline{X}_1, \overline{Tc}_1 \rangle \sqsupseteq_\mathcal{M} \langle \overline{X}_2, \overline{Tc}_2 \rangle$ iff, after computing the temporal transitive closure, there exists a variable mapping $\mathcal{M} : \overline{X}_1 \mapsto \overline{X}_2$ such that for all $X_{1_i}, Y_{1_j} \in \overline{X}_1$ exist $X_{2_m}, Y_{2_n} \in \overline{X}_2$ which satisfy $(X_{1_i} \ (R_{1_{i,j}}) \ Y_{1_j}) \sqsupseteq_\mathcal{M} (X_{2_m} \ (R_{2_{m,n}}) \ Y_{2_n})$.*

*Proof.* "$\Leftarrow$" Since from definition 6.12 $(X_{1_i} \ (R_{1_{i,j}}) \ Y_{1_j}) \sqsupseteq_\mathcal{M} (X_{2_m} \ (R_{2_{m,n}}) \ Y_{2_n})$ implies that $(R_{1_{i,j}})^\mathcal{E} \supseteq (R_{2_{m,n}})^\mathcal{E}$ for every $\mathcal{E}$, then, from the definition of interpretation of a temporal constraint network, it is easy to see that each assignment of variables $\mathcal{V}$ in the interpretation of $\langle \overline{X}_2, \overline{Tc}_2 \rangle$ is also an assignment in the interpretation of $\langle \mathcal{M}(\overline{X}_1), \mathcal{M}(\overline{Tc}_1) \rangle$.
"$\Rightarrow$" Suppose that one is not able to find such a mapping; then, by hypothesis, for each possible variable mapping there exists some $i, j$ such that $R_{1_{i,j}}$ is not a superset of $R_{2_{m,n}}$.

Since, by assumption, the temporal constraint networks are minimal, the temporal relation $R_{2_{m,n}}$ cannot be further restricted. So, for each variable mapping and each temporal interpretation $\mathcal{E}$, we can build an assignment $\mathcal{V}^*$ such that $\langle \mathcal{V}^*(X_{2_m}), \mathcal{V}^*(X_{2_n}) \rangle \in (R_{2_{m,n}})^{\mathcal{E}}$ while $\langle \mathcal{V}^*(X_{1_i}), \mathcal{V}^*(X_{1_j}) \rangle \notin (R_{1_{i,j}})^{\mathcal{E}}$. Now, we can extend the assignment $\mathcal{V}^*$ in such a way that $\mathcal{V}^* \in (\langle \overline{X}_2, \overline{Tc}_2 \rangle)^{\mathcal{E}}$ while $\mathcal{V}^* \notin (\langle \mathcal{M}(\overline{X}_1), \mathcal{M}(\overline{Tc}_1) \rangle)^{\mathcal{E}}$. This contradicts the assumption that $\langle \overline{X}_1, \overline{Tc}_1 \rangle \sqsupseteq_{\mathcal{M}} \langle \overline{X}_2, \overline{Tc}_2 \rangle$. □

**Definition 6.16 (S-mapping)** *An* s-mapping *from a conceptual temporal constraint network $\langle \overline{X}_1, \overline{Tc}_1, \overline{Q@X}_1 \rangle$ to a conceptual temporal constraint network $\langle \overline{X}_2, \overline{Tc}_2, \overline{Q@X}_2 \rangle$ is a variable mapping $\mathcal{S} : \overline{X}_1 \mapsto \overline{X}_2$ such that the non-temporal concept labeling each node in $\overline{X}_1$ subsumes the non-temporal concept labeling the corresponding node in $\mathcal{S}(\overline{X}_1)$, and $\langle \overline{X}_1, \overline{Tc}_1 \rangle \sqsupseteq_{\mathcal{S}} \langle \overline{X}_2, \overline{Tc}_2 \rangle$.*

The algorithm for checking subsumption between temporal concept expressions reduces the subsumer and the subsumee in essential graph form, then it looks for an s-mapping between the essential graphs by exhaustive search. To prove the completeness of the overall subsumption procedure it will be showed that the introduction of LCS's preserves the subsumption. A model-theoretic characterization of the LCS will be given for showing this property. Let's start to build an Herbrand model for an $\mathcal{F}$ concept. Let $C'(x)$ denote the first order formula corresponding to a concept $C$ (see proposition 2.1), while the functionality of features can be expressed with a set of formulæ $\boldsymbol{F}$. By syntax induction it easy to show that $C'(x)$ is an existentially quantified formula with one free variable. Moreover, the matrices of such formula is a conjunction of positive predicates. $\boldsymbol{F} \cup \{C'(x)\}$ is logically equivalent to $\boldsymbol{F} \cup \{C''(x)\}$ where the functionality axioms allow to map every subformula $\bigwedge_y \exists y. F_f(x, y)$ into $\exists! y. F_f(x, y)$. Then $C''(x)$ is such that all the existential quantifiers in $C'(x)$ (which come from the first order conversion of features) are replaced by $\exists!$ quantifiers. Now, $\boldsymbol{F} \cup \{C'''(a)\}$ – where $a$ is a constant substituting the free variable $x$ and $C'''(a)$ is obtained by skolemizing the $\exists!$ quantified variables – is a set of definite Horn clauses.

**Definition 6.17 (Herbrand model)** *Let $C$ be an $\mathcal{F}$ concept expression. Then we define its* Minimal Herbrand Model $\mathcal{H}_C$ *as the* Minimal Herbrand Model *of the above mentioned set of definite Horn clauses $\boldsymbol{F} \cup \{C'''(a)\}$.*

**Lemma 6.18 ($\mathcal{F}$ concept subsumption)** *Let $C, D$ be $\mathcal{F}$ concept expressions, and $\mathcal{H}_C, \mathcal{H}_D$ their minimal Herbrand models obtained by skolemizing the first order set $\boldsymbol{F} \cup \{C'''(a), D'''(a)\}$. Then, $C \sqsubseteq D$ iff $\mathcal{H}_D \subseteq \mathcal{H}_C$.*

*Proof.* $C \sqsubseteq D$ iff $\boldsymbol{F} \cup \{C'(x)\} \models D'(x)$, iff $\boldsymbol{F} \cup \{C''(x)\} \models D''(x)$, where $C''$ and $D''$ are obtained by applying the functionality axioms to the set $\{C'(x), D'(x)\}$ (i.e., unifying the variables in the functional predicates) and then replacing all the existential quantifiers by $\exists!$ quantifiers. Now, $C'''(x)$ and $D'''(x)$ are obtained by skolemizing the $\exists!$ quantified variables in the following way: let $C''(x) = \exists! y_1, \ldots, y_n \phi(x, y_1, \ldots, y_n)$ and let $D''(x) = \exists! y_1, \ldots, y_k, z_1, \ldots, z_m \psi(x, y_1, \ldots, y_k, z_1, \ldots, z_m)$, with $0 \leq k \leq n$, then skolemize the formula: $\sigma = \exists! y_1, \ldots, y_n, z_1, \ldots, z_m \phi(x, y_1, \ldots, y_n) \wedge \psi(x, y_1, \ldots, y_k, z_1, \ldots, z_m)$, and let $\sigma^0(x)$ indicate its skolemized form. Then, $C'''(x) = \phi^0(x)$ and $D'''(x) = \psi^0(x)$. Now, since every existential quantification in $C''(x), D''(x)$ was of type $\forall \exists!$ then the thesis is true iff $\boldsymbol{F} \cup \{C'''(a)\} \models D'''(a)$, where $a$ is a constant substituting the free variable $x$ (see (van Dalen, 1994)). Now, as showed by lemma 6.17, both $C'''(a)$ and $D'''(a)$ have minimal Herbrand models $\mathcal{H}_\mathcal{C}, \mathcal{H}_\mathcal{D}$ that verify the lemma hypothesis. Then, $\boldsymbol{F} \cup \{C'''(a)\} \models D'''(a)$ iff $\mathcal{H}_\mathcal{D} \subseteq \mathcal{H}_\mathcal{C}$. □

We are now able to give a model-theoretic characterization of the LCS that will be crucial to prove the subsumption-preserving property.

**Lemma 6.19 (LCS model property)** *Let $Q_1, \ldots, Q_n$ be $\mathcal{F}$ concept expressions, and $\mathcal{H}_{Q_1}, \ldots, \mathcal{H}_{Q_n}$ their minimal Herbrand models obtained by skolemizing the first order set $\boldsymbol{F} \cup \{Q_1'''(a), \ldots, Q_n'''(a)\}$. Then, $Q = \text{LCS}\{Q_1, \ldots, Q_n\}$ iff $\mathcal{H}_\mathcal{Q} = \mathcal{H}_{Q_1} \cap \ldots \cap \mathcal{H}_{Q_n}$.*

*Proof.* First of all, let show that $\mathcal{H}_\mathcal{Q}$ is the minimal Herbrand model of a concept $Q$ in the language $\mathcal{F}$. Every $\mathcal{H}_{Q_i}$ can be seen as a rooted directed acyclic graph where nodes are labelled with (possible empty) set of atomic concepts and arcs with atomic features while equality constraints between nodes correspond to features agreement. Whithout loss of generality let us consider the case where $\mathcal{H}_\mathcal{Q} = \mathcal{H}_{Q_1} \cap \mathcal{H}_{Q_2}$. It is sufficient to show that $\mathcal{H}_\mathcal{Q}$ is a rooted directed acyclic graph. Let $a$ be the root of $\mathcal{H}_{Q_1}, \mathcal{H}_{Q_2}$, then will be proved by induction that if $F_i(a_{i-1}, a_i) \in \mathcal{H}_\mathcal{Q}$ (where $F_i$ is the first order translation of a feature, $a_{i-1}, a_i$ are obtained as a result of the skolemization process, and $a_0 = a$) then $\{F_1(a, a_1), \ldots, F_i(a_{i-1}, a_i)\} \subseteq \mathcal{H}_\mathcal{Q}$. The case $i = 1$ is trivial. Let $i > 1$. Now, $F_i(a_{i-1}, a_i) \in \mathcal{H}_\mathcal{Q}$ iff $F_i(a_{i-1}, a_i) \in \mathcal{H}_{Q_1} \cap \mathcal{H}_{Q_2}$. But $a_{i-1}$ is uniquely defined by the skolem function $f_{F_{i-1}}$ (where, the function symbols $f_{F_i}$ are newly generated for each feature $F_i$ by the skolemization process). Then, $F_i(a_{i-1}, a_i) \in \mathcal{H}_{Q_1} \cap \mathcal{H}_{Q_2}$ iff $F_i(a_{i-1}, f_{F_{i-1}}(a_i)) \in \mathcal{H}_{Q_1} \cap \mathcal{H}_{Q_2}$ iff $F_{i-1}(a_{i-2}, f_{F_{i-1}}(a_i)) \in \mathcal{H}_{Q_1} \cap \mathcal{H}_{Q_2}$. Then the thesis is true by induction.

Let us now prove the "⇐" direction. Suppose by absurd that there is an $\mathcal{F}$ concept $C$ such that: $Q_1 \sqsubseteq C \wedge Q_2 \sqsubseteq C \wedge C \sqsubset Q$. Then, $Q_1 \sqsubseteq C$ iff $\mathcal{H}_\mathcal{C} \subseteq \mathcal{H}_{Q_1}$, and $Q_2 \sqsubseteq C$, iff $\mathcal{H}_\mathcal{C} \subseteq \mathcal{H}_{Q_2}$. But then $\mathcal{H}_\mathcal{C} \subseteq \mathcal{H}_{Q_1} \cap \mathcal{H}_{Q_2}$, i.e., $\mathcal{H}_\mathcal{C} \subseteq \mathcal{H}_\mathcal{Q}$. Then $Q \sqsubseteq C$ which contradicts the hypothesis.

The "⇒" direction can be proved with analogous considerations. □

**Proposition 6.20 (LCS subsumption-preserving property)** *Let $A, B, C, D$ be $\mathcal{F}$ concepts, then $A \sqcap (B \sqcup C) \sqsubseteq D$ iff $A \sqcap \text{LCS}\{B, C\} \sqsubseteq D$.*

*Proof.* $A \sqcap (B \sqcup C) \sqsubseteq D$ iff $A \sqcap B \sqsubseteq D$ and $A \sqcap C \sqsubseteq D$. Now, $A \sqcap B \sqsubseteq D$ iff $\boldsymbol{F} \cup \{A'''(a), B'''(a)\} \models D'''(a)$ iff $\mathcal{H}_\mathcal{A} \cup \mathcal{H}_\mathcal{B} \models D'''(a)$ iff $\mathcal{H}_\mathcal{D} \subseteq \mathcal{H}_\mathcal{A} \cup \mathcal{H}_\mathcal{B}$. For the same reasons, $A \sqcap C \sqsubseteq D$ iff $\mathcal{H}_\mathcal{D} \subseteq \mathcal{H}_\mathcal{A} \cup \mathcal{H}_\mathcal{C}$. But then, $\mathcal{H}_\mathcal{D} \subseteq \mathcal{H}_\mathcal{A} \cup \mathcal{H}_\mathcal{B}$ and $\mathcal{H}_\mathcal{D} \subseteq \mathcal{H}_\mathcal{A} \cup \mathcal{H}_\mathcal{C}$, i.e., $\mathcal{H}_\mathcal{D} \subseteq \mathcal{H}_\mathcal{A} \cup (\mathcal{H}_\mathcal{B} \cap \mathcal{H}_\mathcal{C})$, i.e., $\mathcal{H}_\mathcal{D} \subseteq \mathcal{H}_\mathcal{A} \cup \mathcal{H}_{\text{LCS}\{B,C\}}$. But, $\mathcal{H}_\mathcal{D} \subseteq \mathcal{H}_\mathcal{A} \cup \mathcal{H}_{\text{LCS}\{B,C\}}$ iff $A \sqcap \text{LCS}\{B, C\} \sqsubseteq D$. □

The following theorem provides a sound and complete procedure to compute subsumption. The completeness proof takes into account that the temporal structure is dense and unbounded. This allows us to introduce any new node to a conceptual temporal constraint network without changing its meaning. Remember that, for each of these redundant nodes, time-invariant information holds.

**Theorem 6.21 ($\mathcal{TL}$-$\mathcal{F}$ concept subsumption)** *A concept $C_1$ subsumes a concept $C_2$ iff there exists an s-mapping from the essential graph of $C_1$ to the essential graph of $C_2$.*

*Proof.* Let $T_1 = \langle \overline{X}_1, \overline{Tc}_1, \overline{Q@X_1} \rangle$ be the essential graph of $C_1$, and $T_2 = \langle \overline{X}_2, \overline{Tc}_2, \overline{Q@X_2} \rangle$ be the essential graph of $C_2$.

"$\Leftarrow$" (Soundness). Follows from the fact that the essential graph form is logically equivalent to the starting concept, and from the soundness of the procedures for computing both the TCN subsumption (proposition 6.15) and the subsumption between non-temporal concepts (Cohen et al., 1992).

"$\Rightarrow$" (Completeness). Suppose that such an s-mapping does not exist. Two main cases can be distinguished.

i) There is not a mapping $\mathcal{M}$ such that $\langle \overline{X}_1, \overline{Tc}_1 \rangle \sqsupseteq_{\mathcal{M}} \langle \overline{X}_2, \overline{Tc}_2 \rangle$. By adding redundant nodes to $T_2$, an equivalent conceptual temporal constraint network $T_2^* = \langle \overline{X}_2^*, \overline{Tc}_2^*, \overline{Q@X_2^*} \rangle$ may be obtained. Let us consider such an extended network in a way that there exists a variable mapping $\mathcal{M}^*$ such that $\langle \overline{X}_1, \overline{Tc}_1 \rangle \sqsupseteq_{\mathcal{M}^*} \langle \overline{X}_2^*, \overline{Tc}_2^* \rangle$. Now, for all possible $\mathcal{M}^*$, there is a node $X_{1_i} \in \overline{X}_1$ such that $\mathcal{M}^*(X_{1_i}) = X_{2_j}$ with $X_{2_j} \notin \overline{X}_2$. Now, $Q_{1_i} \not\sqsupseteq_{\mathcal{F}} Q_{2_j}$, since $X_{2_j}$ cannot coincide with other nodes in $\overline{X}_2$ neither can have a covering otherwise the hypothesis that the mapping $\mathcal{M}$ does not exist would be contradicted. Then from proposition 6.7 $Q_{2_j}$ is in a time-invariant node, whereas $Q_{1_i}$ is not since $T_1$ is an essential graph. Then, although the construction of $\mathcal{M}^*$ allows for the existence of a unique $\mathcal{V}^3$ for both networks (follows from proposition 6.15), it is possible to build an instance of $T_2$ that is not an instance of $T_1$.

ii) For each possible mapping $\mathcal{M}$ such that $\langle \overline{X}_1, \overline{Tc}_1 \rangle \sqsupseteq_{\mathcal{M}} \langle \overline{X}_2, \overline{Tc}_2 \rangle$ there will be always two nodes $X_{1_i}$ and $X_{2_j}$ such that $\mathcal{M}(X_{1_i}) = X_{2_j}$ and $Q_{1_i} \not\sqsupseteq_{\mathcal{F}} Q_{2_j}$. Now, the concept expression $Q_{2_j}$ cannot be refined (looking for a subsumption relationship with $Q_{1_i}$) by adding to it an $\mathcal{F}$ concept since from proposition 6.7 this would change the overall interpretation. On the other hand, the LCS introduction – which would substitute the more specific concept disjunction implicitly presents because of a node covering – is a subsumption-invariant concept substitution, as showed by lemma 6.20.

Both cases contradict the assumption that $T_1$ subsumes $T_2$. □

### 6.2.1 COMPLEXITY OF SUBSUMPTION

Now it is shown that checking subsumption between $\mathcal{TL}$-$\mathcal{F}$ concept expressions in the essential graph form is an NP-complete problem. Therefore, a polynomial reduction from the NP-complete problem of deciding whether a graph contains an isomorphic subgraph is presented. It is then shown that the subsumption computation, as proposed in theorem 6.21, can be done by a non-deterministic algorithm that takes polynomial time in the size of the concepts involved. First of all let us consider the complexity of computing subsumption between non-temporal concepts.

**Lemma 6.22 ($\mathcal{F}$ subsumpion complexity)** *Let $C, D$ be $\mathcal{F}$ concept expressions that can contain LCS's. Then, checking whether $C \sqsubseteq_{\mathcal{F}} D$ takes polynomial time.*

*Proof.* See (Cohen et al., 1992). □

Here the problem of *subgraph isomorphism* is briefly recalled. Given two graphs, $G_1 = (V_1, E_1)$ and $G_2 = (V_2, E_2)$, $G_1$ contains a subgraph isomorphic to $G_2$ if there exists a subset of the vertices $V' \subseteq V_1$ and a subset of the edges $E' \subseteq E_1$ such that $|V'|=|V_2|$, $|E'|=|E_2|$, and there exists a one-to-one function $f: V_2 \mapsto V'$ satisfying $\{u,v\} \in E_2$ iff $\{f(u), f(v)\} \in E'$.

---

3. Since subsumption is computed with respect to a fixed evaluation time, $\mathcal{V}$ maps the different occurrences of $\sharp$ to the same interval; this justifies the choice that $\mathcal{M}(\sharp) = \sharp$.

Given a graph $G = (V, E)$, with $V = \{v_1, \ldots, v_n\}$ associate a temporal concept expression: $C \doteq \Diamond(v_1, \ldots, v_n) \ldots (v_i \ (\mathsf{b},\mathsf{a}) \ v_j) \ldots (A@v_1 \sqcap \ldots \sqcap A@v_n)$, where $A$ is an atomic concept and $\{v_i, v_j\} \in E$. This transformation allows us to prove that the problem of subgraph isomorphism can be reduced to the subsumption of temporal concepts.

**Proposition 6.23** *Given two graphs $G_1$ and $G_2$, $G_1$ contains a subgraph isomorphic to $G_2$ iff $C_2 \sqsupseteq C_1$, where $C_1$ and $C_2$ are the corresponding temporal concepts expressions.*

*Proof.* A temporal network with edges labeled only with the (before∨after) relation is always consistent, minimal and non-directed[4] (Gerevini & Schubert, 1994). Then, each temporal concept is in the essential graph form. Now the proof easily follows since, every time $G_2$ is an isomorphic subgraph of $G_1$ the one-to-one function $f$ is also an s-mapping from $C_2$ to $C_1$, and it is true that $C_2 \sqsupseteq C_1$. On the other hand, the s-mapping that gives rise to the subsumption is also the one-to-one isomorphism from $G_2$ to $G_1$. □

**Theorem 6.24 (NP-hardness)** *Concept subsumption between $\mathcal{TL}$-$\mathcal{F}$ concept expressions in normal form is an NP-hard problem.*

*Proof.* Follows from proposition 6.23 and the reduction being clearly polynomial. □

Now the NP-completeness is proven.

**Theorem 6.25 (NP-completeness)** *Concept subsumption between $\mathcal{TL}$-$\mathcal{F}$ concept expressions in normal form is an NP-complete problem.*

*Proof.* To prove NP-completeness it is necessary to show that the proposed calculus can be solved by a nondeterministic algorithm that takes polynomial time. Now, given two temporal concepts, $T_1$ and $T_2$, in their essential graph form, let $|\overline{X}_1| = N_1$ and $|\overline{X}_2| = N_2$. Then, to check whether $T_1 \sqsupseteq T_2$, the algorithm guesses one of the $N_2^{N_1}$ variable mapping from $T_1$ to $T_2$ and verifies whether it is an s-mapping, too. This last step can be done in deterministic polynomial time since, given a mapping $\mathcal{M}$, it is possible to determine whether $\langle \overline{X}_1, \overline{Tc}_1 \rangle \sqsupseteq_{\mathcal{M}} \langle \overline{X}_2, \overline{Tc}_2 \rangle$ by checking at most $N_1(N_1-1)/2$ edges looking for subsumption between the corresponding temporal relations (solved by a set inclusion procedure); while the $N_1$ non-temporal concept subsumptions can be computed in polynomial time. □

## 7. Extending the Propositional Part of the Language

The propositional part of the temporal language can be extended to have a more powerful, but still decidable, Description Logic. It is possible either to add full disjunction, both at the temporal and non-temporal levels ($\mathcal{TLU}$-$\mathcal{FU}$), or to have a propositionally complete language at the non-temporal level only ($\mathcal{TL}$-$\mathcal{ALCF}$).

Please note that in these languages it is not possible to express full negation, and in particular the negation of the existential temporal quantifier. This is crucial, and it

---

4. If $(v_i \ (\mathsf{b},\mathsf{a}) \ v_j)$ then $(v_j \ (\mathsf{b},\mathsf{a}) \ v_i)$, too.

A Temporal Description Logic for Reasoning about Actions and Plans

$$
\begin{aligned}
(C \sqcup D)@X &\rightarrow C@X \sqcup D@X \\
p : (C \sqcup D) &\rightarrow p : C \sqcup p : D \\
(C_1 \sqcup C_2) \sqcap D &\rightarrow (C_1 \sqcap D) \sqcup (C_2 \sqcap D) \\
\diamond(\overline{X})\,\overline{\mathit{tc}}.\,(C \sqcup D) &\rightarrow \diamond(\overline{X})\,\overline{\mathit{tc}}.\,C \sqcup \diamond(\overline{X})\,\overline{\mathit{tc}}.\,D
\end{aligned}
$$

Figure 13: Rewrite rules for computing the disjunctive form.

makes the difference with other logic-based approaches (Schmiedel, 1990; Bettini, 1997; Halpern & Shoham, 1991). The dual of $\diamond$ (i.e., the universal temporal quantifier $\square$) makes the satisfiability problem – and the subsumption – for propositionally complete languages undecidable in the most interesting temporal structures (Halpern & Shoham, 1991; Venema, 1990; Bettini, 1993). For the representation of actions and plans in the context of plan recognition, the universal temporal quantifier is not strictly necessary. This limitation makes these languages decidable, with nice computational properties, and capable of supporting other kinds of useful extensions. The examples shown throughout the paper may serve as a partial validation of the claim. Section 8.1 proposes the introduction of a limited universal temporal quantification that maintains decidability of subsumption.

### 7.1 Disjunctive Concepts: $\mathcal{TLU}$-$\mathcal{FU}$

The language $\mathcal{TLU}$-$\mathcal{FU}$ adds to the basic language $\mathcal{TL}$-$\mathcal{F}$ the disjunction operator – with the usual semantics – both at the temporal and non-temporal levels:

$$
\begin{aligned}
C, D &\rightarrow \mathcal{TL} \mid C \sqcup D \quad (\mathcal{TLU}) \\
E, F &\rightarrow \mathcal{F} \mid E \sqcup F \quad (\mathcal{FU})
\end{aligned}
$$

Before showing how to modify the calculus to check subsumption, let us begin with a clarifying example. The gain in expressivity allows us to describe the alternative realizations that a given plan may have. Let us consider a scenario with a robot moving in an empty room that can move only either horizontally or vertically. Let's call `Rect-Move` that which involves a simple sequence of the two basic moving actions. Then, to describe a `Rect-Move` plan we can make use of the disjunction operator:

$$
\begin{aligned}
\texttt{Rect-Move} \doteq\ &\diamond(x\ y)\,(\sharp\ \text{m}\ x)(x\ \text{m}\ y).\,\big(\texttt{Hor-Move}@x \sqcap \texttt{Ver-Move}@y\big) \sqcup \\
&\diamond(x\ y)\,(\sharp\ \text{m}\ x)(x\ \text{m}\ y).\,\big(\texttt{Ver-Move}@x \sqcap \texttt{Hor-Move}@y\big)
\end{aligned}
$$

#### 7.1.1 The Calculus for $\mathcal{TLU}$-$\mathcal{FU}$

NORMAL FORM

In computing subsumption, a normal form for concepts is needed. The normalization procedure is similar to that reported in Section 6.1. Let us start by reducing each concept expression into an equivalent *disjunctive concept* of the form:

$$
(\diamond(\overline{X}_1)\,\overline{\mathit{tc}}_1.\,G_1) \sqcup \cdots \sqcup (\diamond(\overline{X}_n)\,\overline{\mathit{tc}}_n.\,G_n)\ \sqcup Q_1 \sqcup \cdots \sqcup Q_m
$$

where $G_i$ are conjunctions of concepts of the form $Q_{i_k}@X_{i_k}$, and each $Q$ does not contain neither temporal information, nor disjunctions, i.e., it is an element of the language $\mathcal{F}$.

**Proposition 7.1 (Equivalence of disjunctive form)** *Every concept $C$ can be reduced into an equivalent disjunctive form (df $C$), by exhaustively applying the set of rewrite rules of Figure 13 in addition to the rules introduced in Figure 11.*

It is now possible to compute the *completed disjunctive normal form* (cdnf $C$). Each disjunct of such normal form has some interesting properties, which are crucial for the proof of the theorem 7.4 on concept subsumption: temporal constraints are always explicit, i.e., any two intervals are related by a basic temporal relation; there is no disjunction, either implicit or explicit, neither in the conceptual part nor in the temporal part, i.e., it is a $\mathcal{TL}$-$\mathcal{F}$ concept; the information in each node is independent of the information in the other nodes and it does not contain time-invariant (i.e., redundant) nodes.

**Definition 7.2 (Completed disjunctive normal form)** *Given a concept in* disjunctive form, *the* completed disjunctive normal form *is obtained by applying the following rewrite rules to each disjunct:*

- *(*Temporal completion*) The rules of definition 6.5 are applied to each disjunct with the exclusion of the* covering *step, which is replaced by the $\sqcup$-introduction step. If a disjunct is unsatisfiable – i.e., the temporal constraint network associated with it is inconsistent – then eliminate it.*

- *(*Essential form*) The rules of definition 6.8 are applied to each disjunct.*

- *($\sqcup$ introduction) Reduce to concepts containing only basic temporal relationships:*
  $$\Diamond(\overline{X})(X_1\ (R,S)\ X_2)\overline{\mathit{Tc}}.C \quad \to \quad \Diamond(\overline{X})(X_1\ R\ X_2)\overline{\mathit{Tc}}.C \ \sqcup\ \Diamond(\overline{X})(X_1\ S\ X_2)\overline{\mathit{Tc}}.C$$

**Proposition 7.3 (Equivalence of CDNF)** *Every concept expression can be reduced into an equivalent completed disjunctive normal concept.*

SUBSUMPTION

The theorem 7.4 reduces subsumption between CDNF concepts into subsumption of disjunction-free concepts, such that the results of theorem 6.21 can be applied. The following theorem gives a terminating, sound, and complete subsumption calculus for $\mathcal{TLU}$-$\mathcal{FU}$.

**Theorem 7.4 ($\mathcal{TLU}$-$\mathcal{FU}$ concept subsumption)** *Let $C = C_1 \sqcup \cdots \sqcup C_m$ and $D = D_1 \sqcup \cdots \sqcup D_n$ be $\mathcal{TLU}$-$\mathcal{FU}$ concepts in CDNF. Then, $C \sqsubseteq D$ if and only if $\forall i \exists j.\ C_i \sqsubseteq D_j$.*

*Proof.* Since it is easy to show that $C_1 \sqcup \ldots \sqcup C_n \sqsubseteq D$ iff $\forall i.C_i \sqsubseteq D$ we need only to prove the restricted thesis: $C_i \sqsubseteq D_1 \sqcup \cdots \sqcup D_n$ iff $C_i \sqsubseteq D_1 \vee \ldots \vee C_i \sqsubseteq D_n$. Every concept expression in CDNF corresponds to an existential quantified formula with two free variables. Moreover, the matrices of such formulæ are conjunctions of positive predicates. Let us denote the formula corresponding to a concept $C$ as $C'(t,x)$. Now, the restricted thesis holds iff it is true that $F \cup \{C_i'''(a,b)\} \models D_1'''(a,b) \vee D_2'''(a,b)$. Now, let $\mathcal{H}_\mathcal{B}$ the minimal Herbrand model of $F \cup \{C_i'''(a,b)\}$. Then, $F \cup \{C_i'''(a,b)\} \models D_1'''(a,b) \vee D_2'''(a,b)$ iff $\mathcal{H}_\mathcal{B} \models D_1'''(a,b) \vee D_2'''(a,b)$. Since we are talking of a single model, $D_1'''(a,b) \vee D_2'''(a,b)$ is valid in $\mathcal{H}_\mathcal{B}$ if and only if either $D_1'''(a,b)$ or $D_2'''(a,b)$ is valid in $\mathcal{H}_\mathcal{B}$. This proves the theorem.[5] □

---

5. The proof of this theorem comes from an idea of Werner Nutt.

As a consequence of the theorems 6.25, 7.4 the following complexity result holds.

**Theorem 7.5 ($\mathcal{TLU}$-$\mathcal{FU}$ subsumption complexity)** *Concept subsumption between $\mathcal{TLU}$-$\mathcal{FU}$ concept expressions in normal form is an NP-complete problem.*

### 7.2 A Propositionally Complete Language: $\mathcal{TL}$-$\mathcal{ALCF}$

$\mathcal{TL}$-$\mathcal{ALCF}$ uses the propositionally complete Description Logic $\mathcal{ALCF}$ (Hollunder & Nutt, 1990) for non-temporal concepts by changing the syntax rules for $\mathcal{TL}$-$\mathcal{F}$ in the following way:

$$E, F \rightarrow \mathcal{FU} \mid \bot \mid \neg E \mid p \uparrow q \mid p\uparrow \mid \forall P.E \mid \exists P.E \quad (\mathcal{ALCF})$$

The interpretation functions are extended to take into account roles:

$$P^{\mathcal{I}} \subseteq \mathcal{T}_{<}^{\star} \times \Delta^{\mathcal{I}} \times \Delta^{\mathcal{I}}$$
$$P_t^{\mathcal{I}} = \hat{P}_t \subseteq \Delta^{\mathcal{I}} \times \Delta^{\mathcal{I}} \mid \forall a, b. \ \langle a, b \rangle \in \hat{P}_t \leftrightarrow \langle t, a, b \rangle \in P^{\mathcal{I}}$$

As seen in Section 2, $\mathcal{ALCF}$ adds to $\mathcal{F}$ full negation – thus introducing disagreement ($p \uparrow q$) and undefinedness ($p\uparrow$) for features, and role quantification ($\forall P.E, \exists P.E$).

As an example of the expressive power gained, let us refine the description of the world states involved in the Stack action (see Section 5.2). Suppose that a block is described by saying that it has LATERAL-SIDEs (role) and BOTTOM- and TOP-SIDEs (features). Then, the property of being clear could be represented as follows:

Clear-Block $\doteq$ Block $\sqcap$ $\forall$LATERAL-SIDE.Clear $\sqcap$ TOP-SIDE : HAS-ABOVE $\uparrow$

which says that, in order to be clear, each LATERAL-SIDE has to be clear and nothing has to be over the TOP-SIDE. Now, the situation in which a block involved in a Stack action is on top of another one is reformulated with the following concept expression:

($\star$OBJECT1∘TOP-SIDE∘HAS-ABOVE $\downarrow$ $\star$OBJECT2)

Furthermore, given the above definition of Clear-Blocks, it can be derived that:

($\star$OBJECT1∘TOP-SIDE∘HAS-ABOVE $\downarrow$ $\star$OBJECT2) $\sqsubseteq$ ($\star$OBJECT1 : $\neg$Clear-Block)

i.e., an object, having another object on top of it, is no more a clear object.

In $\mathcal{TL}$-$\mathcal{ALCF}$ it is possible to describe states with some form of incomplete knowledge by exploiting the disjunction among non-temporal concepts. For example, let us say that the agent of an action can be either a human being or a machine: $\star$AGENT.(Person$\sqcup$Robot).

#### 7.2.1 THE CALCULUS FOR $\mathcal{TL}$-$\mathcal{ALCF}$

This Section presents a calculus for deciding subsumption between temporal concepts in the Description Logic $\mathcal{TL}$-$\mathcal{ALCF}$. Again, the calculus is based on the idea of separating the inference on the temporal part from the inference on the Description Logic part ("$\sqsubseteq_{\mathcal{ALCF}}$"), and adopting standard procedures developed in the two areas.

NORMAL FORM

Once more, the subsumption calculus is based on a normalization procedure. The first step reduces a concept expression into an equivalent *existential form* – $\Diamond(\overline{X}) \ \overline{\mathcal{T}c}. \ (Q_0 \sqcap Q_1@X_1 \sqcap \ldots \sqcap Q_n@X_n)$ – by applying the rewrite rules of Figure 11 augmented with the

$$
\begin{array}{rcl}
\neg \top & \to & \bot \\
\neg \bot & \to & \top \\
\neg (C \sqcap D) & \to & \neg C \sqcup \neg D \\
\neg (C \sqcup D) & \to & \neg C \sqcap \neg D \\
\neg \neg C & \to & C \\
\neg \forall P.C & \to & \exists P.\neg C \\
\neg \exists P.C & \to & \forall P.\neg C \\
\neg f : C & \to & f \uparrow \sqcup f : \neg C \\
\neg p : C & \to & f \uparrow \sqcup f : (\neg q : C) \quad \text{if } p = f \circ q \\
\neg p \downarrow q & \to & p \uparrow \sqcup q \uparrow \sqcup p \uparrow q \\
\neg p \uparrow q & \to & p \uparrow \sqcup q \uparrow \sqcup p \downarrow q \\
(f \circ p) \uparrow & \to & f \uparrow \sqcup f : (p \uparrow)
\end{array}
$$

*Note:* By $f$ we denote both an atomic feature and an atomic parametric feature.

Figure 14: Rewrite rules to transform an arbitrary concept into a simple concept.

rule: $p : (q_1 \uparrow q_2) \to p \circ q_1 \uparrow p \circ q_2$. Each $Q$ is a non-temporal concept, i.e., it is an element of the language $\mathcal{ALCF}$.

In the following normalization step there will be a need to verify concept satisfiability for non-temporal concept expressions. An $\mathcal{ALCF}$ concept $E$ is unsatisfiable iff $E \sqsubseteq_{\mathcal{ALCF}} \bot$. Algorithms for checking satisfiability and subsumption of concepts terms in $\mathcal{ALCF}$ are well known (Hollunder & Nutt, 1990).

**Definition 7.6 (Completed existential form)** *The* temporal completion *of a concept in existential form – the* Completed Existential Form, *CEF – is obtained by sequentially applying the following steps:*

- *(closure, collapsing, covering) As reported in definition 6.5. As for the* covering, *translate the concept expression $Q_y$ applying the rewrite rule: $Q_y \to Q_y \sqcap_{i=1\ldots n}(Q_{i_1} \sqcup \ldots \sqcup Q_{i_m})$.*

- *(parameter introduction) This requires two phases.*

  1. *Each $Q$ is translated in* disjunctive normal form. *First the simple form[6] is obtained by transforming each $Q$ following the rewrite rules reported in Figure 14. The* disjunctive normal form *is then obtained by rewriting each $Q$ – which is now in simple form – using the following rules, which correspond to the first order rules for computing the disjunctive normal form of logical formulæ:*

  $$
  \begin{array}{rcl}
  (C_1 \sqcup C_2) \sqcap D & \to & (C_1 \sqcap D) \sqcup (C_2 \sqcap D) \\
  p : (C \sqcup D) & \to & p : C \sqcup p : D
  \end{array}
  $$

---

6. A *simple concept* contains only complements of the form $\neg A$, where $A$ is a primitive concept, and no sub-concepts of the form $p\uparrow$, where $p$ is not an atomic (parametric) feature – this corresponds to a first order logical formula in negation normal form.

$$
\begin{array}{rcl}
\star g_1 \circ \ldots \circ \star g_n \, [\circ \, f \, [\circ \, p]] : C & \to & \star g_1 \circ \ldots \circ \star g_n : \top. \\
\star g_1 \circ \ldots \circ \star g_n \, [\circ \, f \, [\circ \, p]] \downarrow g \, [\circ \, q] & \to & \star g_1 \circ \ldots \circ \star g_n : \top. \\
\star g_1 \circ \ldots \circ \star g_n \downarrow \star f_1 \circ \ldots \circ \star f_m & \to & \star g_1 \circ \ldots \circ \star g_n \downarrow \star f_1 \circ \ldots \circ \star f_m. \\
\star g_1 \circ \ldots \circ \star g_n \circ g \, [\circ \, p] \downarrow \star f_1 \circ \ldots \circ \star f_m \, [\circ \, f \, [\circ \, q]] & \to & \star g_1 \circ \ldots \circ \star g_n : \top \sqcap \\
& & \star f_1 \circ \ldots \circ \star f_m : \top. \\
\star g_1 \circ \ldots \circ \star g_n \uparrow \star f_1 \circ \ldots \circ \star f_m & \to & \star g_1 \circ \ldots \circ \star g_n \uparrow \star f_1 \circ \ldots \circ \star f_m. \\
\star g \uparrow & \to & \star g \uparrow. \\
\star g_1 \circ \ldots \circ \star g_n : (\star g_{n+1} \uparrow) & \to & \star g_1 \circ \ldots \circ \star g_n : (\star g_{n+1} \uparrow). \\
\star g_1 \circ \ldots \circ \star g_n \, [\circ \, f \, [\circ \, p]] \uparrow g \, [\circ \, q] & \to & \star g_1 \circ \ldots \circ \star g_n : \top. \\
\star g_1 \circ \ldots \circ \star g_n \circ g \, [\circ \, p] \uparrow \star f_1 \circ \ldots \circ \star f_m \, [\circ \, f \, [\circ \, q]] & \to & \star g_1 \circ \ldots \circ \star g_n : \top \sqcap \\
& & \star f_1 \circ \ldots \circ \star f_m : \top.
\end{array}
$$

Figure 15: Rewrite rules that compute the parameter introduction step.

2. For each $Q_j = E_{j_1} \sqcup \ldots \sqcup E_{j_n}$, on compute its time-invariant part (let us indicate this particular concept expression as $\tilde{Q}_j$). This gives $\tilde{Q}_j$ by computing for each disjunct $E_{j_i}$ in $Q_j$ its time-invariant information $\tilde{E}_{j_i}$. If $E_{j_i} \sqsubseteq_{\mathcal{ALCF}} \bot$, then $\tilde{E}_{j_i} = \bot$. Otherwise, rewrite every conjunct in $E_{j_i}$ as showed in Figure 15, while the conjuncts not considered there are rewrote to $\top$. Now, unless there is an $\tilde{E}_{j_i} = \top$, $\tilde{Q}_j = \tilde{E}_{j_1} \sqcup \ldots \sqcup \tilde{E}_{j_n}$ must be conjunctively added to all the other nodes.

**Proposition 7.7 (Equivalence of CEF)** *Every concept in existential form can be reduced into an equivalent completed existential concept.*

As for the $\mathcal{TL}$-$\mathcal{F}$ case, both *covering* and *parameter introduction* can be computed independently. As a consequence of the above normalization phase, the proposition 6.7 (node independence) is now true for $\mathcal{TL}$-$\mathcal{ALCF}$ concepts in CEF. Observe that, to obtain a CEF concept, the steps of the normalization procedure require the computation of the transitive closure of the temporal relations – which is an NP-complete problem (van Beek & Cohen, 1990) – and the computation of $\mathcal{ALCF}$ subsumption – which is a PSPACE-complete problem (Hollunder & Nutt, 1990).

Before the presentation of the last normalization phase, which will eliminate redundant nodes, it is now possible to check whether a concept expression is satisfiable.

**Proposition 7.8 (Concept satisfiability)** *A $\mathcal{TL}$-$\mathcal{ALCF}$ concept in CEF, $\langle \overline{X}, \overline{Tc}, \overline{Q@X} \rangle$, is satisfiable (with the proviso that the temporal constraints are satisfiable) if and only if the non-temporal concepts labeling each node in $\overline{X}$ are satisfiable. Checking satisfiability of a $\mathcal{TL}$-$\mathcal{ALCF}$ concept in CEF is a PSPACE-complete problem.*

*Proof.* Is a direct consequence of the node independence established by proposition 6.7, which is true also for $\mathcal{TL}$-$\mathcal{ALCF}$ concepts in CEF. □

The normalization procedure now goes on by rewriting unsatisfiable concepts to $\bot$ and then computing the *essential graph form* for satisfiable concepts. This last phase is more complex than for the other temporal languages considered in this paper essentially because $\mathcal{ALCF}$ can express the $\top$ concept by means of a concept expression (e.g., $\top = A \sqcup \neg A$). From this consideration it follows that in $\mathcal{TL}\text{-}\mathcal{ALCF}$ a redundant node can be derived from a complex concept expression (e.g., both $A \sqcup \neg A$, and $\star g : A \sqcup \star g : \neg A$ are redundant nodes). The key idea is that all the time-invariant information is present in the $\sharp$ node thanks to the CEF. Thus it is needed only to extract this information from the $\sharp$ node by computing the disjunctive normal form of $Q_\sharp$, applying the $\tilde{\cdot}$ translation, and then testing whether $\tilde{Q}_\sharp \sqsubseteq_{\mathcal{ALCF}} Q_i$, for a given node $x_i$.

**Definition 7.9 (Essential graph)** *The subgraph of the CEF of a $\mathcal{TL}\text{-}\mathcal{ALCF}$ conceptual temporal constraint network $T = \langle \overline{X}, \overline{Tc}, \overline{Q@X} \rangle$ obtained by deleting the nodes $x_i$ such that $\tilde{Q}_\sharp \sqsubseteq_{\mathcal{ALCF}} Q_i$ – with the exception of the $\sharp$ node – is called* essential graph *of $T$: (ess $T$).*

**Proposition 7.10 (Equivalence of essential graph)** *Every CEF concept can be reduced into an equivalent essential graph form (and, obviously, every concept can be reduced into an equivalent essential graph form).*

SUBSUMPTION

The overall normalization procedure reduces the subsumption problem in $\mathcal{TL}\text{-}\mathcal{ALCF}$ to the subsumption between $\mathcal{ALCF}$ concepts.

**Theorem 7.11 ($\mathcal{TL}\text{-}\mathcal{ALCF}$ concept subsumption)** *A concept $C_1$ subsumes a concept $C_2$ if and only if there exists an s-mapping from the essential graph of $C_1$ to the essential graph of $C_2$.*

The above theorem gives a sound and complete algorithm for computing subsumption between $\mathcal{TL}\text{-}\mathcal{ALCF}$ concepts (the proof is the same as the one for theorem 6.21). The subsumption problem is now PSPACE-hard, since satisfiability and subsumption for $\mathcal{ALCF}$ concepts were proven to be PSPACE-complete (Hollunder & Nutt, 1990).

## 8. Extending the Expressivity for States

The following suggests how to extend the basic language to cope with important issues in the representation of states. (i) Homogeneity allows us to consider properties of the world – peculiar to states – which remain true in each subinterval of the interval in which they hold. (ii) Persistence guarantees that a state holding as an effect of an action continues to hold unless there is no evidence of its falsity at some time. An approach to the *frame* problem is then presented, showing a possible solution to one of the most (in)famous problems in AI literature. The following subsections shall be interested more in semantically characterizing actions and states than on computational properties. The extensions proposed now to the temporal languages are for having a full fledged *Description Logic for time and action*.

### 8.1 Homogeneity

In the temporal literature *homogeneity* characterizes the temporal behavior of world states: when a state holds over an interval of time $t$, it also holds over subintervals of $t$. Thus, if a block is on the table for a whole day, one can conclude that it is also on the table in the morning. On the other hand, actions are not necessarily homogeneous. In the linguistic literature a difference is made between *activity* and *performance* verbs. The distinction comes out in the fact that activity verbs do have sub-events that are denoted by the same verb, whereas performance verbs do not. Generally, activity verbs represent ongoing events, for example *to eat* and *to run*, and can be described as homogeneous predicates; whereas performance verbs represent events with a well defined granularity in time, such as *to prepare spaghetti*. Performance verbs are an example of anti-homogeneous events: if they occur over an interval of time $t$, then they do not occur over a subinterval of $t$, as they would not yet be completed.

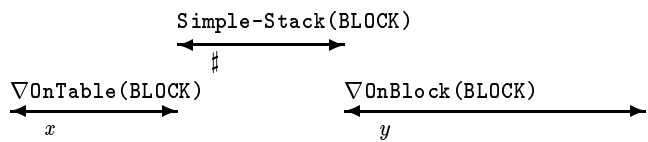

Figure 16: Temporal dependencies in the definition of the `Simple-Stack` action.

The language is extended by introducing the *Homogeneity* operator:

$$C, D \quad \rightarrow \quad \nabla C \qquad \text{(homogeneous concept)}$$

The semantics of homogeneous concepts is easily given in terms of the semantics of the temporal universal quantifier: $\nabla C \equiv \Box x \ (x \ (=,\mathsf{s},\mathsf{d},\mathsf{f}) \ \sharp).\ C@x$. This means that $\nabla C$ is an homogeneous concept if and only if when it holds at an interval it remains true at each subinterval. In particular, $\Box x$ universally qualifies the temporal variable $x$, while the temporal constraint $(x \ (=,\mathsf{s},\mathsf{d},\mathsf{f}) \ \sharp)$ imposes that $x$ is a generic interval contained in $\sharp$. Moreover, it is always true that $\nabla C \sqsubseteq C$, i.e., $\nabla C$ is a more specific concept than $C$.

Let us consider as an example a more accurate definition of the `Basic-Stack` action (see Section 4.1.1):

$$\texttt{Simple-Stack} \ \doteq \ \Diamond(x\,y)(x\ \mathsf{m}\ \sharp)(\sharp\ \mathsf{m}\ y).\ \big((\star\texttt{BLOCK} : \nabla\texttt{OnTable})@x \sqcap$$
$$(\star\texttt{BLOCK} : \nabla\texttt{OnBlock})@y\big)$$

Figure 16 shows the temporal dependencies of the intervals in which the `Simple-Stack` holds. The difference with the `Basic-Stack` action is the use of the homogeneity operator. In fact, since the predicates `OnTable` and `OnBlock` denote states, their homogeneity should be explicitly declared. The assertion `Simple-Stack`$(i, a)$ says that $a$ is an individual action of type `Simple-Stack` occurred at interval $i$. Moreover, the same assertion implies that $a$ is related to a $\star$`BLOCK`, say $b$, which is of type `OnTable` at some interval $j$ – meeting $i$ – *and* at all intervals included in $j$, while it is of type `OnBlock` at another interval $l$ – met by $i$ – *and* at all intervals included in $l$:

$$\texttt{Simple-Stack}(i,a) \Longrightarrow \exists b.\ \star\texttt{BLOCK}(a,b) \land \exists\, j, l.\ \mathsf{m}(j,i) \land \mathsf{m}(i,l) \land$$
$$\forall \hat{\jmath}, \hat{l}.\ (=,\mathsf{s},\mathsf{d},\mathsf{f})(\hat{\jmath}, j) \land (=,\mathsf{s},\mathsf{d},\mathsf{f})(\hat{l}, l) \rightarrow$$
$$\texttt{OnTable}(\hat{\jmath}, b) \land \texttt{OnBlock}(\hat{l}, b).$$

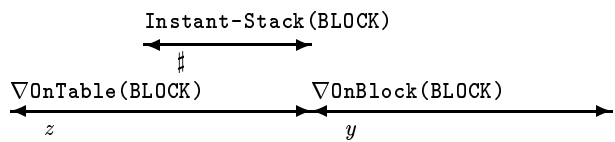

Figure 17: Temporal dependencies in the definition of the Instant-Stack action.

Note that the Simple-Stack action subsumes the Instant-Stack action, whose temporal dependencies are depicted in Figure 17:

Instant-Stack $\doteq \Diamond(z\,y)(\sharp\,\text{f}\,z)(\sharp\,\text{m}\,y).\,((\star\text{BLOCK}:\nabla\text{OnTable})@z \sqcap$
$(\star\text{BLOCK}:\nabla\text{OnBlock})@y)$

Subsumption holds because the class of intervals – obtained by homogeneity of the state OnTable as defined in the Simple-Stack action – including $x$ and all its subintervals is a subset of the class of intervals over which the block is known to be on the table, according to the definition of Instant-Stack – this latter class includes all the subintervals of $z$.

If the Instant-Stack action had been defined without the $\nabla$ operator, then it would not specialize any more the Simple-Stack action. In fact, according to such a weaker definition of Instant-Stack, specifying that the object is on the table at $z$ does not imply that the object is on the table at subintervals of $z$; in particular, it is not possible to deduce any more that the object is on the table at $x$ and its subintervals, as specified in the definition of Simple-Stack action. Moreover, the *weak* Instant-Stack action type would not specialize the *weak* Simple-Stack action type – i.e., Basic-Stack – too. Thus, homogeneity helps us to define states and actions in a more accurate way, such that important inferences are captured.

As seen above, the definition of homogeneity makes use of universal temporal quantification. Remember that subsumption in a propositionally complete Description Logic with both existential and universal temporal quantification is undecidable and it is still an open problem if it becomes decidable in absence of negation (Bettini, 1993). The homogeneity operator is a restricted form of universal quantification. An even more restricted form interests us here, where the concept $C$ in $\nabla C$ does not contain any other temporal operator (called *simple homogeneous concept*). The expressiveness of the resulting logic is enough, for example, to correctly represent the homogeneous nature of states. In (Artale, Bettini, & Franconi, 1994) an algorithm to compute subsumption in $\mathcal{TL}\text{-}\mathcal{F}$ augmented with the homogeneity operator is proposed. Even if a formal proof is still not available, good arguments are discussed to conjecture its completeness. This would also prove decidability of this logic and of the corresponding modal logics.

### 8.2 Persistence

This Section shows how our framework can be successfully extended in a general way to cope with *inertial* properties. In the basic temporal language, a property holding, say, as a post-condition of an action at a *certain* interval, is not guaranteed to hold anymore at other included or subsequent intervals. This is the reason why we propose an extended formalism, in which *states* can be represented as homogeneous and persistent concepts. As a motivation for introducing the possibility of representing persistent properties in the language, this Section considers how to solve the *frame problem*, and in particular the famous example of the Yale Turkey Shooting Scenario (Sandewall, 1994; Allen & Ferguson, 1994), formerly known as the *Yale Shooting Problem*.

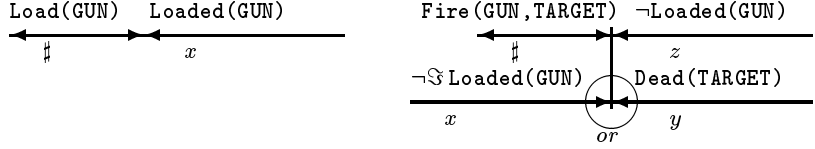

Figure 18: Definitions of the actions Load and Fire.

An inertia operator "$\Im$" is introduced here. Intuitively, $\Im C$ is currently true if it was true at a preceding interval – say $i$ – and there is no evidence of the falsity of $C$ at any interval between the current one and $i$. Thus, the property of an individual of being of type $C$ *persists* over time, unless a contradiction arises.

The formalization of the inertia operator makes use of the epistemic operator **K** (Donini, Lenzerini, Nardi, Schaerf, & Nutt, 1992), in which **K**$C$ denotes the set of individuals *known* to be instances of the concept $C$[7].

**Definition 8.1 (Inertia)** $\Im C(j, a)$ iff
$\exists i.\ start(i) \leq start(j)\ \wedge\ C(i, a)\ \wedge$
  $\forall h.\ start(h) \geq end(i)\ \wedge\ end(h) \leq end(j)\ \rightarrow\ \neg\mathbf{K}\neg C(h, a).$

where *start* and *end* are two functions giving respectively the starting and the ending point of an interval – conditions on endpoints are simpler and more readable than their equivalents on interval relations; $\neg\mathbf{K}\neg C(h, a)$ means that it is not known that $a$ is not of type $C$ at interval $h$. Furthermore, the following relation holds: $\forall a, j.\ C(j, a) \rightarrow \Im C(j, a)$; i.e., $\Im C$ subsumes $C$. The above definition can be captured by a temporal language equipped with the epistemic operator – **K** – and the homogeneity operator – $\nabla$:

$\Im C \equiv C \sqcup \Diamond(x\ y)\ (x\ (\mathsf{b},\mathsf{m},\mathsf{o},\mathsf{fi},\mathsf{di})\ \sharp)(x\ (\mathsf{s},\mathsf{si})\ y)(y\ \mathsf{fi}\ \sharp).(C@x \sqcap \nabla(\neg\mathbf{K}\neg C)@y)$

Two action types are defined, Load – with the parameter ⋆GUN – and Fire – with the parameters ⋆GUN and ⋆TARGET (Figure 18):

Load $\doteq$ $\Diamond x\ (\sharp\ \mathsf{m}\ x).$ ⋆GUN : Loaded@$x$

Fire $\doteq$ $\Diamond(x\ y\ z)\ (\sharp\ \mathsf{f}\ x)(\sharp\ \mathsf{m}\ y)(\sharp\ \mathsf{m}\ z).$
        (⋆GUN : $\neg\Im$Loaded@$x \sqcup$ ⋆TARGET : Dead@$y$) $\sqcap$ ⋆GUN : $\neg$Loaded@$z$

The action Load describes loading a gun. The action Fire describes firing the gun against a target: effects of firing are that the gun becomes unloaded and either the target is dead

---

7. An *epistemic interpretation* is a pair $(\mathcal{I}, \mathcal{W})$ in which $\mathcal{I}$ is an interpretation and $\mathcal{W}$ is a set of interpretations such that $(\mathbf{K}C)^{\mathcal{I}, \mathcal{W}} = \bigcap_{\mathcal{J} \in \mathcal{W}}(C^{\mathcal{J}, \mathcal{W}})$.

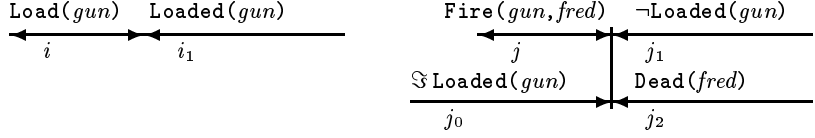

Figure 19: Actions instances in the Yale Shooting Problem.

or the gun was not loaded – possibly by inertia – before firing. The Yale Shooting Problem considers the situation described by the following set of assertions (ABox):

Load($i$, *load-action*), ⋆GUN(*load-action*, *gun*), a($j, i$), Fire($j$, *fire-action*),
⋆GUN(*fire-action*, *gun*), ⋆TARGET(*fire-action*, *fred*).

i.e., at the beginning the *gun* is loaded; then, the action of firing the *gun* against the target *fred* is performed. According to the semantics of the language, logical consequences of the knowledge base $\Sigma$ are:

$\Sigma \models \exists i_1.\ \mathsf{m}(i, i_1) \wedge \mathtt{Loaded}(i_1, gun)$

$\Sigma \models \exists j_1.\ \mathsf{m}(j, j_1) \wedge \neg\mathtt{Loaded}(j_1, gun)$

$\Sigma \models \exists j_0.\ \mathsf{f}(j, j_0) \wedge \Im\mathtt{Loaded}(j_0, gun)$

$\Sigma \models \exists j_2.\ \mathsf{m}(j, j_2) \wedge \mathtt{Dead}(j_1, fred)$.

i.e., (see also Figure 19) (i) the Load action makes the *gun* loaded; (ii) the Fire action makes the *gun* unloaded *at the end*; (iii) since there is no evidence to the contrary, the *gun* is still loaded at $j_0$ by inertia; (iv) since the gun is not unloaded at $j_0$, the target *fred* must be dead.

Since the inertia operator is useful to describe the behavior of *properties*, which are characterized as homogeneous concepts, a simple way of representing persistence in the context of homogeneous concepts is proposed.

**Proposition 8.2** *Let $P$ be a property – i.e., $P \doteq \nabla P'$ is an homogeneous concept – and $\Sigma$ a knowledge base such that $\Sigma \not\models P(j, a)$. $\Im P(j, a)$ is true in $\Sigma$ – i.e., $\Sigma \models \Im P(j, a)$ – if and only if two intervals $i, k$ exist such that: $\Sigma \models (start(i) \leq start(j) \wedge P(i, a))$ and $\Sigma \cup \{\mathsf{s}(i, k),\ \mathsf{f}(j, k),\ P(k, a)\}$ is satisfiable.*

*Proof.* The entailment test verifies the first part of the definition of inertia, while the satisfiability test verifies that, between the interval at which the system knows that the individual $a$ belongs to $P - i$ – and the interval at which $P(a)$ is deduced by inertia – $j$ – does not exist an interval $h$ at which the system knows that $P(a)$ is false. Indeed, such interval $h$ would be related to the interval $k$ by the relation in and since it is supposed that $P$ is homogeneous, the knowledge base with $\neg P(h, a) \wedge P(k, a) \wedge \mathsf{in}(h, k)$ would be inconsistent. □

The deduction $P(j, a) \to \Im P(j, a)$ can be obtained as a particular case of the above stated proposition.

## 9. Related Works

The original formalism devised by Allen (1991) forms, in its very basis, the foundation for our work. It is a predicate logic in which interval temporal networks can be introduced, properties can be asserted to *hold* over intervals, and events can be said to *occur* at intervals. His approach is very general, but it suffers from problems related to the semantic formalization of the predicates `hold` and `occur` (Blackburn, 1992). Moreover, computational properties of the formalism are not analyzed. The study of this latter aspect was, on the contrary, our main concern.

In the Description Logic literature, other approaches for representing and reasoning with time and action were proposed. In the beginning the approaches based on an explicit notion of time are surveyed, and then the STRIPS-like approaches are considered. This Section ends by illustrating some of the approaches devoted to temporally extend the situation calculus.

Bettini (1997) suggests a variable-free extension with both existential and universal temporal quantification. He gives undecidability results for a class of temporal languages – resorting to the undecidability results of Halpern and Shoham's temporal logic – and investigates approximated reasoning algorithms. Basically, he extends the $\mathcal{ALCN}$ description logics with the existential and universal temporal quantifiers, but, unlike our formalism, explicit interval variables are not allowed. The temporal quantification makes use of a set of temporal constraints on two implicit intervals: the reference interval and the current one. In this framework, the concept of `Mortal` can be defined as:

`Mortal` $\doteq$ `LivingBeing` $\sqcap$ $\Diamond$(after). (**not** `LivingBeing`)

Schild (1993) proposes the embedding of *point*-based tense operators in a propositionally closed Description Logic. He proved that satisfiability in $\mathcal{ALCT}$, the point-based temporal extension of $\mathcal{ALC}$, interpreted on a linear, unbounded and discrete temporal structure, is PSPACE-complete. His ideas were applied by (Fischer, 1992; Neuwirth, 1993) in the BACK system. Note that a point-based temporal ontology is unable to express all the variety of relations between intervals.

Baader and Laux (1995) integrate modal operators for time and belief in a terminological system looking for an adequate semantics for the resulting combined language. The major point in this paper is the possibility of using modal operators not only inside concept expressions but also in front of concept definitions and assertions. The following example shows the notion of `Happy-father`, where different modalities interact:

[BEL-JOHN](`Happy-father` $\doteq$ $\exists$MARRIED-TO.(`Woman` $\sqcap$ [BEL-JOHN]`Pretty`) $\sqcap$
$\langle$future$\rangle\forall$CHILD.`Graduate`)

In this case, it is *John*'s belief that a `Happy-father` is someone married to a woman believed to be pretty by *John*, and whose children will be graduates sometime in the future. The semantics has a Kripke-style: each modal operator is interpreted as an accessibility relation on a set of possible worlds, while the domain of objects is split into (possible) different domain objects, each one depending on a given world. This latter choice captures the case of different definitions for the same concept – such as [BEL-JOHN]($A \doteq B$) and [BEL-PETER]($A \doteq C$) – since the two formulæ are evaluated in different worlds. The main restriction is that all the modal operators do not satisfy any specific axioms for belief or time. On the other hand, the language is provided with a complete and terminating algorithm that should serve, as the authors propose, "...as a basis for satisfiability algorithms for more complex languages".

There are Description Logics intended to represent and reasoning about actions following the STRIPS tradition. Heinsohn, Kudenko, Nebel and Profitlich (1992) describe the RAT system, used in the WIP project at the German Research Center for AI (DFKI). They use a Description Logic to represent both the world states and atomic actions. A second formalism is added to compose actions in plans and to reason about simple temporal relationships. No explicit temporal constraints can be expressed in the language. RAT actions are defined by the change of the world state they cause, and they are instantaneous as in the STRIPS-like systems, while plans are linear sequences of actions. The most important service offered by RAT is the *simulated execution* of part of a plan, checking if a given plan is *feasible* and, if so, computing the global pre- and post-conditions. The feasibility test is similar to the usual consistency check for a concept description: they *temporally project* the pre- and post-conditions of individual actions composing the plan, respectively backward and forward. If this does not lead to an inconsistent initial, final or intermediate state, the plan is feasible and the global pre- and post-conditions are determined as a side effect.

Devanbu and Litman (1991, 1996) describe the CLASP system, a *plan-based* knowledge representation system extending the notion of subsumption and classification to plans, to build an efficient information retrieval system. In particular, CLASP was used to represent plan-like knowledge in the domain of telephone switching software by extending the use of the software information system LASSIE (Devanbu, Brachman, Selfridge, & Ballard, 1991). CLASP is designed for representing and reasoning about large collections of plan descriptions, using a language able to express temporal, conditional and looping operators. Following the STRIPS tradition, plan descriptions are built starting from states and actions, both represented by using the CLASSIC (Brachman, McGuiness, Patel-Schneider, Resnick, & Borgida, 1991) terminological language. Since plans constructing operators correspond to regular expressions, algorithms for subsumption integrate work in automata theory with work in concept subsumption. The temporal expressive power of this system can capture to sequences, disjunction and iterations of actions and each action is instantaneous. Furthermore, state descriptions are restricted to a simple conjunction of primitive CLASSIC concepts. Like RAT, CLASP checks if an instantiated plan is well formed, i.e., the specified sequence of individual actions are able to transform the given initial state into the goal state by using the STRIPS rules.

We end up by reporting on the efforts made by researchers in the situation calculus field to overcome the strict sequential perspective inherent to this framework. Recent works enrich the original framework to represent properties and actions having different truth values depending not only on the situation but also on time. The work of Reiter (1996), moving from the results showed by Pinto (1994) and by Ternovskaia (1994), provides a new axiomatization of the situation calculus able to capture concurrent actions, properties with continuous changes, and natural exogenous actions – those under nature's control. The notion of *fluent* – which models properties of the world – and *situation* are maintained. Each action is instantaneous and responsible for changing the actual situation to the subsequent one. Concurrent actions are simply sets of instantaneous actions that must be coherent, i.e., the action's collection must be non empty and all the actions occur at the same time. Pinto (1994) and Reiter (1996) introduce the time dimension essentially to capture both the occurrence of the natural actions, due to known laws of physics – i.e., the ball bouncing at times prescribed by motion's equations – and the dynamic behavior of physical objects – i.e., the position of a falling ball. This is realized by introducing a time argument for each action function, while properties of the world are divided into two different classes: classical fluents that hold or do not hold throughout situations, and *continuous parameters* that may change their value during the time spanned by the given situation.

More devoted to have a situation calculus with a time interval ontology is the work of Ternovskaia (1994). In order to describe *processes* – i.e., actions extended in time – she introduces durationless actions that initiate and terminate those processes. As a matter of fact, processes become fluents, with instantaneous events – *Start(Fluent)* and *Finish(Fluent)* – which respectively make true or false the corresponding fluent, and with persistence assumptions that make the fluent true during the interval. For example, in a blocks world the *picking-up* process is treated as a fluent with *Start(picking-up(x))* and *Finish(picking-up(x))* instantaneous actions that enable or falsify the *picking-up* fluent.

## 10. Conclusions

The main objective of this paper was the design of a class of logical formalisms for uniformly representing time, actions and plans. According to this framework, an action has a duration in time, it can have parameters, which are the ties with the temporal evolution of the world, and it is possibly associated over time with other actions. A model-theoretic semantics including both a temporal and an object domain was developed, for giving both a meaning to the language formulæ and a well founded definition of the various reasoning services, allowing us to prove soundness and completeness of the corresponding algorithms. The peculiar computational properties of this logic make it an effective representation and reasoning tool for plan recognition purposes. An action taxonomy based on subsumption can be set up, and it can play the role of a plan library for plan retrieval tasks.

This paper contributes to exploration of the decidable realm of interval-based temporal extensions of Description Logics. It presented complete procedures for subsumption reasoning with $\mathcal{TL}$-$\mathcal{F}$, $\mathcal{TLU}$-$\mathcal{FU}$ and $\mathcal{TL}$-$\mathcal{ALCF}$. In addition, the subsumption problem for $\mathcal{TL}$-$\mathcal{F}$ was proven an NP-complete problem. The subsumption procedures are based on an interpretation preserving transformation that operates a separation between the temporal and the non-temporal parts of the formalism. Thus, the calculus can adopt distinct standard procedures developed in the Description Logics community and in the temporal constraints community. To obtain decidable languages the key idea was to restrict the temporal expressivity by eliminating the universal quantification on temporal variables. While a propositionally complete Description Logic with both existential and universal temporal quantification is undecidable, it is still an open problem if it becomes decidable in absence of negation. With the introduction of the homogeneity operator investigation of the impact of a restricted form of temporal universal quantification in the language $\mathcal{TL}$-$\mathcal{F}$ was begun.

Several extensions were proposed to the basic temporal language. With the possibility to specify homogeneous predicates the temporal behavior of world states can be described in a more natural way, while the introduction of the non-monotonic inertial operator gives rise to some forms of temporal prediction. Another extension – not considered in this paper – deals with the possibility of relating an action to more elementary actions, *decomposing* it in partially ordered steps (Artale & Franconi, 1995). This kind of reasoning is found in hierarchical planners like NONLIN (Tate, 1977), SIPE (Wilkins, 1988) and FORBIN (Dean, Firby, & Miller, 1990).

## Acknowledgements

This paper is a substantial extension and revision of (Artale & Franconi, 1994). The work was partially supported by the Italian National Research Council (CNR) project "Ontologic and Linguistic Tools for Conceptual Modeling", and by the "Foundations of Data Warehouse Quality" ($DWQ$) European ESPRIT IV Long Term Research (LTR) Project 22469. The first author wishes to acknowledge also LADSEB-CNR of Padova and the University of Firenze for having supported part of his work. Some of the work carried on for this paper was done while the second author was working at ITC-irst, Trento. This work owes a lot to our colleagues Claudio Bettini and Alfonso Gerevini, for having introduced us many years ago to the *temporal maze*. Special thanks to Achille C. Varzi, for taking time to review the technical details of the paper and for his insightful comments on the philosophy of events, and to Fausto Giunchiglia, for useful discussions and feedback. Thanks to Paolo Bresciani, Nicola Guarino, Eugenia Ternovskaia and Andrea Schaerf for enlightening comments on earlier drafts of the paper. Werner Nutt and Luciano Serafini helped us to have a deeper insight into logic. We would also like to thank Carsten Lutz for the helpful discussions we had with him about temporal representations. Many anonymous referees checked out many errors of previous versions of the paper. All the errors of the paper are, of course, our own.